\documentclass[journal,twoside,web]{ieeecolor}

\usepackage{etoolbox}
\makeatletter
\@ifundefined{color@begingroup}%
{\let\color@begingroup\relax
\let\color@endgroup\relax}{}%
\def\fix@ieeecolor@hbox#1{%
\hbox{\color@begingroup#1\color@endgroup}}
\patchcmd\@makecaption{\hbox}{\fix@ieeecolor@hbox}{}{\FAILED}
\patchcmd\@makecaption{\hbox}{\fix@ieeecolor@hbox}{}{\FAILED}

\usepackage{tmi}
\usepackage{cite}
\usepackage{amsmath,amssymb,amsfonts}
\usepackage{algorithmic}
\usepackage{graphicx}
\usepackage{textcomp}
\usepackage[utf8]{inputenc} 
\usepackage[T1]{fontenc}    
\usepackage{hyperref}       
\usepackage{url}            
\usepackage{booktabs}       
\usepackage{amsfonts}       
\usepackage{nicefrac}       
\usepackage{microtype}      
\usepackage{graphicx}
\usepackage{lscape}
\usepackage{array}
\usepackage{makecell}
\usepackage{colortbl}
\usepackage{rotating}
\usepackage{multirow}
\usepackage{multicol}
\usepackage{threeparttable}
\usepackage{amsmath,amssymb,graphicx}
\usepackage{bbm}
\usepackage{dsfont}
\usepackage{pifont}
\newcommand{\ie}{\textit{i.e.}, }
\newcommand{\eg}{\textit{e.g.}, }
\newcommand{\etc}{\textit{etc}}

\def\BibTeX{{\rm B\kern-.05em{\sc i\kern-.025em b}\kern-.08em
    T\kern-.1667em\lower.7ex\hbox{E}\kern-.125emX}}
\begin{document}
\title{Contrastive Graph Modeling for Cross-Domain Few-Shot Medical Image Segmentation}
\author{Yuntian Bo, Tao Zhou, Zechao Li, Haofeng Zhang, and Ling Shao,~\IEEEmembership{Fellow,~IEEE} 
 \thanks{Manuscript received 00-00, 2025; revised 00-00, 2025.}
 \thanks{This work was partly supported by the National Natural Science Foundation of China (NSFC) under Grant No. 62371235, and partly by the Key Research and Development Plan of Jiangsu Province (Industry Foresight and Key Core Technology Project) under Grant BE2023008-2. (Corresponding author: Haofeng Zhang.)}
 \thanks{Yuntian Bo, Tao Zhou, Zechao Li and Haofeng Zhang are with the School of Computer Science and Engineering, Nanjing University of Science and Technology, Nanjing, 210094, China. (E-mail:\{yuntian.bo, taozhou, zechao.li, zhanghf\}@njust.edu.cn)}
 \thanks{Ling Shao is with the UCAS-Terminus AI Lab, University of Chinese Academy of Sciences, Beijing, 100190, China. (e-mail: ling.shao@ieee.org) }
 }

\maketitle

\begin{abstract}
Cross-domain few-shot medical image segmentation (CD-FSMIS) offers a promising and data-efficient solution for medical applications where annotations are severely scarce and multimodal analysis is required.
However, existing methods typically filter out domain-specific information to improve generalization, which inadvertently limits cross-domain performance and degrades source-domain accuracy.
To address this, we present Contrastive Graph Modeling (C-Graph), a framework that leverages the structural consistency of medical images as a reliable domain-transferable prior. We represent image features as graphs, with pixels as nodes and semantic affinities as edges. A Structural Prior Graph (SPG) layer is proposed to capture and transfer target-category node dependencies and enable global structure modeling through explicit node interactions. Building upon SPG layers, we introduce a Subgraph Matching Decoding (SMD) mechanism that exploits semantic relations among nodes to guide prediction. Furthermore, we design a Confusion-minimizing Node Contrast (CNC) loss to mitigate node ambiguity and subgraph heterogeneity by contrastively enhancing node discriminability in the graph space.
Our method significantly outperforms prior CD-FSMIS approaches across multiple cross-domain benchmarks, achieving state-of-the-art performance while simultaneously preserving strong segmentation accuracy on the source domain.
Our code is available at \url{https://github.com/primebo1/C-Graph}.
\end{abstract}


\section{Introduction}
Medical image segmentation is fundamental for computer-assisted intervention \cite{MIS_CAI}, yet current mainstream deep learning methods \cite{Unet,nnUnet} still rely heavily on large-scale annotated data. Moreover, the resulting models are typically task-specific and fail to generalize to new concepts.
In recent years, few-shot medical image segmentation (FSMIS) \cite{ssl-alp2020,rp-net} has gained increasing attention due to its potential to reduce data reliance. By exploiting category-agnostic prior knowledge learned from base categories, FSMIS seeks to segment novel anatomical structures with only a few labeled examples. However, current FSMIS approaches naively assume a shared distribution between training and inference data, resulting in substantial performance degradation when deployed in unseen domains \cite{FAMNet}. In contrast, clinical practice frequently involves multiple imaging techniques (\eg CT, MRI) to capture comprehensive patient information, inherently introducing domain shifts that challenge model generalization.

\begin{figure}[!t]
    \centering
    \includegraphics[width=0.99\linewidth]{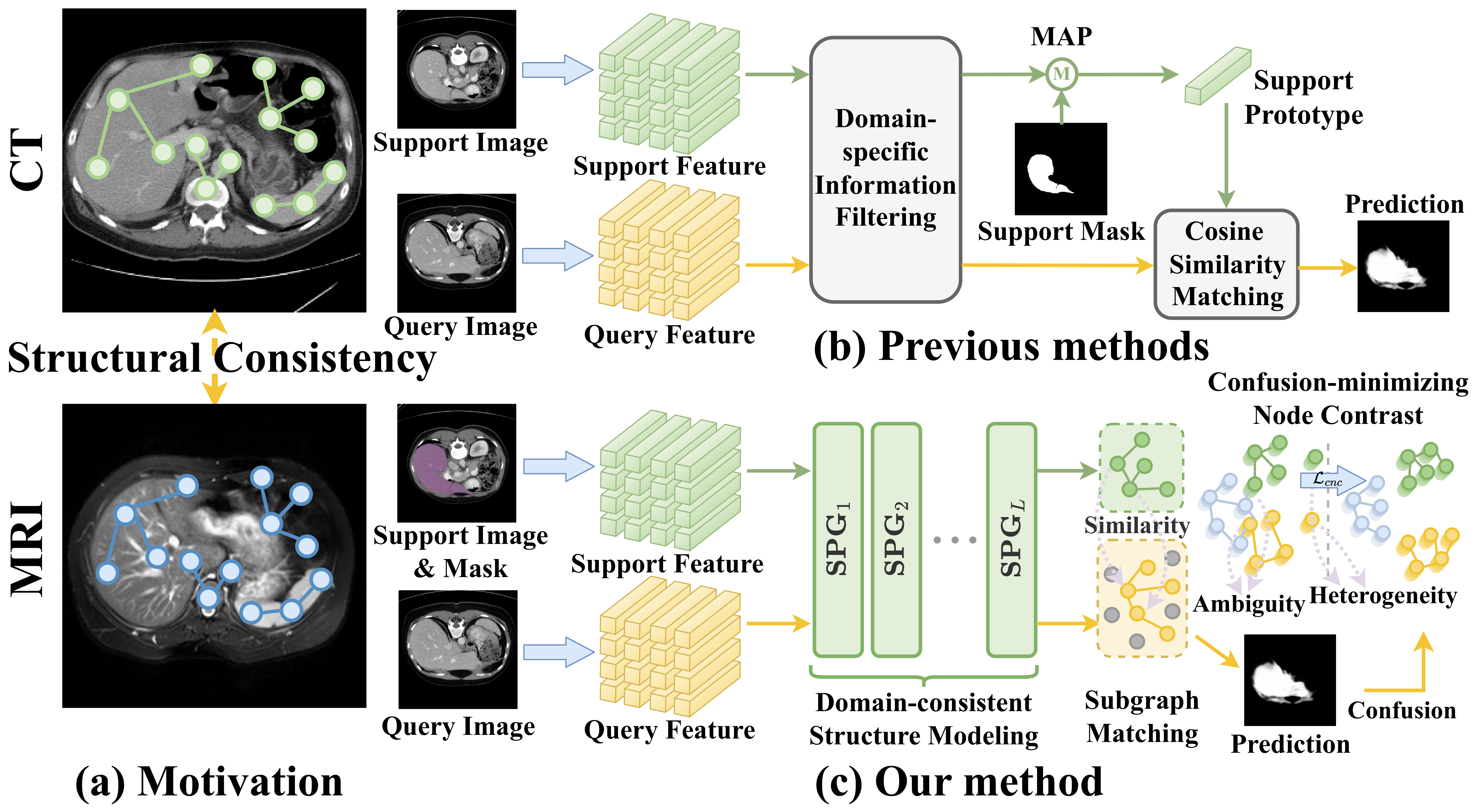}
    \caption{(a) Our motivation. Despite significant appearance shifts, medical images exhibit high structural consistency across domains. (b) Previous methods naively focus on filtering out domain-specific information to improve generalization, while overlooking feature collapse, limiting cross-domain performance and severely degrading source-domain accuracy. (c) Our method models domain-consistent structure using graphs and employs contrastive learning to reduce node semantic confusion, achieving both superior in- and cross-domain performance.}
    \vspace{-2ex}
    \label{fig:motivation}

\end{figure}

The discrepancies above have motivated the formulation of the cross-domain few-shot medical image segmentation (CD-FSMIS) task \cite{ROBUSTEMD,FAMNet}, where model generalization extends beyond segmenting novel categories to handling unseen domains.
Existing CD-FSMIS methods assume domain knowledge is decouplable, thus attempting to filter domain-relevant information for enhanced transferability. For instance, FAMNet \cite{FAMNet} explicitly suppresses domain-specific frequency components to promote generalization. However, this simplistic information filtering inevitably compromises feature integrity, placing an inherent limit on achieving more precise cross-domain segmentation. Moreover, despite improved generalization, we observe a substantial degradation in the source domain, compared to conventional FSMIS methods. This highlights the need to identify more semantically consistent patterns to maximize cross-domain generalization while minimizing source domain performance degradation.

In medical imaging, despite substantial variations across different imaging domains, we observe an inherent structural consistency attributed to anatomical properties and priors of biological tissues and organs. Here, structure is defined as:
\vspace{1ex}

\noindent \textbf{Definition 1 }\textit{(structure). We define structure as the \textbf{semantic relationships} between \textbf{spatial positions} within the feature map, which reflect both categorical attributes and anatomical priors.}

 \vspace{1ex}
\noindent As shown in Fig. \ref{fig:motivation}(a), organ morphology and spatial semantics, defined by the arrangement of pixels, are largely unaffected by modality-induced appearance shifts. This observation motivates us to model such domain-agnostic structure as robust and transferable priors for segmentation in novel target domains.

In light of the foundations above, we are naturally motivated to model structure using graphs. Taking semantic similarity as the edge criterion, a medical image feature map can be transformed into a bottom-up graph: pixel-level features at each spatial position serve as nodes, nodes belonging to the same category form a subgraph, and subgraphs collectively constitute the full graph. However, this modeling approach faces two key challenges: 1) \textit{High node ambiguity.} Semantic ambiguity is prevalent in medical images due to overlapping tissue, fuzzy boundaries, \etc, leading to increased node classification uncertainty during graph optimization. 2) \textit{High subgraph heterogeneity}. Variations in detail emphasis across imaging techniques may introduce considerable heterogeneity among nodes within the same subgraph, thereby reducing its semantic compactness. Together, these sources of semantic confusion degrade the reliability of the modeled graph.

To tackle the aforementioned challenges, we present Contrastive Graph Modeling (C-Graph), a novel framework that reformulates CD-FSMIS as a graph learning problem. At its core, we introduce a simple yet effective Structural Prior Graph (SPG) layer, which serves as the primary building block of C-Graph.
For novel category generalization, SPG layers leverage the support set to capture semantic dependencies among nodes of the target-class subgraph, forming class-level structural knowledge that is interactively transferred and emphasized in the query graph. For unseen domain generalization, SPG layers explicitly enable semantic interactions across spatial positions during training, progressively modeling global domain-agnostic structure.
Moreover, the prevailing practice in (CD-) FSMIS, prototypical matching, treats pixel features as isolated entities, neglecting their interdependencies. To fully leverage the semantic dependencies among nodes modeled by the SPG layers, we propose a novel Subgraph Matching Decoding (SMD) mechanism that explicitly accounts for semantic connectivity among graph nodes during the decoding process, thereby enabling structurally coherent and context-aware segmentation.
Finally, to address the highly confused nodes in graph modeling, we introduce a Confusion-minimizing Node Contrast (CNC) loss, which adaptively exploits semantic edge cues to enhance node discriminability within the graph space, and enforces structure-oriented learning to counteract domain shifts.

Extensive experimental results demonstrate that our method consistently outperforms the previous state-of-the-art (SOTA) approach FAMNet \cite{FAMNet} by large margins, achieving an average improvement of 3.51\% across diverse cross-domain scenarios. Remarkably, it also maintains expert-level performance on the source domain, marking a notable breakthrough in the CD-FSMIS field.
To summarize, we propose five key contributions:
\begin{itemize}
    \item We present Contrastive Graph Modeling (C-Graph), a novel framework that focuses on \emph{structural consistency} in medical images to enhance generalization across domains.
    \item We propose a Structural Prior Graph (SPG) Layer that simultaneously models structure and highlights target query nodes for improved generalization and matching.
    \item We propose a Subgraph Matching Decoding (SMD) mechanism that departs from the prototypical paradigm to effectively leverage the structural patterns learned through the SPG layers for enhanced prediction.
    \item We design a Confusion-minimizing Node Contrast (CNC) loss that reduces semantic confusion by contrastively guiding the model to learn structures regardless of appearance. 
    \item We validate the generalizability of C-Graph on four cross-domain medical datasets. Experimental results demonstrate the SOTA performance of the proposed method. 
\end{itemize}

\begin{figure*}
    \vspace{-1ex}
    \centering
    \includegraphics[width=0.99\linewidth]{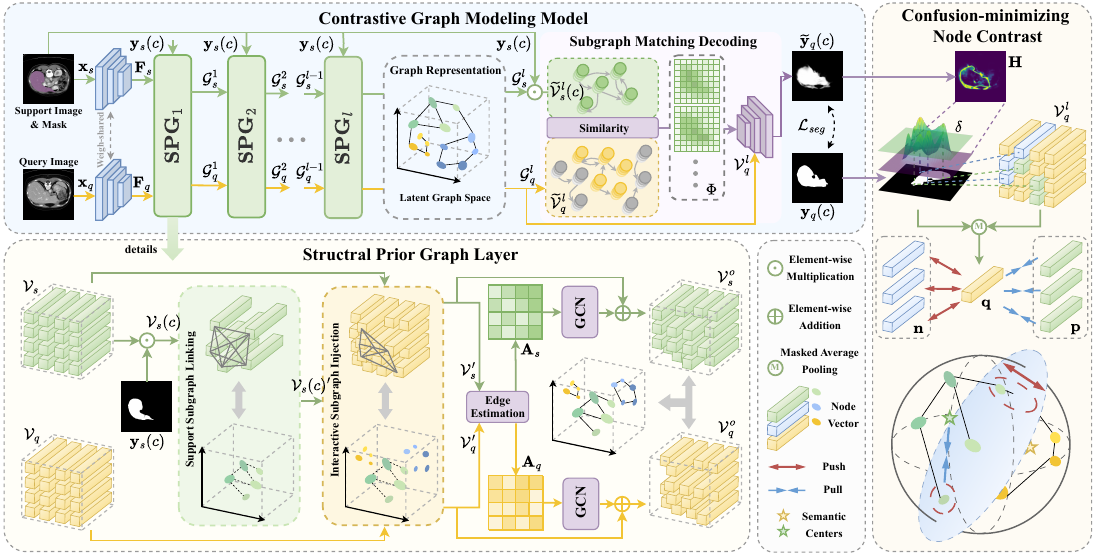}
    \caption{Overview of our proposed method.  Here, $\mathcal{G}^i$ denotes the output graph of the $i$-th SPG layer.}
    \label{fig:overall_arch}
    \vspace{-1ex}
\end{figure*}

\section{Related Works}
\subsection{Few-Shot Medical Image Segmentation}
 The goal of FSMIS is to segment previously unseen categories during inference by providing the model with a few annotated samples as transferable knowledge. Although a few methods \cite{SENET,MRrNet,GCN-DE,AAS-DCL,rap} have explored interactive conditional networks for FSMIS, prototypical network-based approaches have remained the predominant paradigm.
Early works \cite{ssl-alp2020,ssl-alp,ADNET} attempted to mitigate the limited cross-category generalizability resulting from scarce training categories by introducing pseudo-label-based self-supervised strategies, which led to a notable performance breakthrough in FSMIS. More recently, research has shifted towards optimizing prototype acquisition and refinement, matching mechanisms, or both \cite{qnet,PMCR,PGRNET,DSPNet,GMRD,rpt,pami,maup,DIFD}, \eg Huang \textit{et al.} \cite{VQ} improve the clustering and representation of prototypes by reformulating prototype matching as a vector quantization problem. RPT \cite{rpt}  learns query-aligned regional prototypes via a transformer architecture. GMRD \cite{GMRD} generates multiple descriptors to address the insufficiency of a single prototype in capturing the entire class distribution. DSPNet \cite{DSPNet} learns multiple high-fidelity prototypes through self-refinement by attending to detailed semantic and structural information.

Although effective, these methods tend to be highly domain-specific and thus rely heavily on retraining when applied to new imaging domains. However, in realistic scenarios, collecting data and training separate models for each domain is often impractical. As a result, researchers have proposed CD-FSMIS, where models are designed to generalize across domains.

\subsection{Cross-Domain Few-Shot Medical Image Segmentation}
CD-FSMIS was introduced to achieve generalization not only to novel classes but also to unseen imaging domains. Previous approaches primarily focus on suppressing domain-specific information and improving robust feature matching. For example, RobustEMD \cite{ROBUSTEMD} addresses the texture discrepancies of targets across domains and proposes a domain-robust matching mechanism based on Earth Mover’s Distance (EMD), which suppresses texture-sensitive signals and emphasizes class boundary consistency.
FAMNet \cite{FAMNet} identifies significant cross-domain differences concentrated in specific frequency bands, and thus adopts a band-wise matching strategy to selectively suppress domain-specific frequency components and emphasize domain-agnostic ones, thereby simultaneously enhancing model generalizability and facilitating support-query feature de-biasing.

However, previous methods largely neglect the preservation of source-domain performance. In this work, source-domain performance is also considered to maintain the value for clinical applications. Consequently, our approach does not attempt to suppress or constrain domain-specific information, as we argue that such suppression impairs source-domain performance and constrains the upper bound of cross-domain generalization. Instead, we boost CD-FSMIS performance by directly capturing domain-transferable anatomical structures embedded in the image features, achieving strong performance in both in- and cross-domain scenarios.

\subsection{Graph Neural Networks}
Graph Neural Networks (GNNs) have demonstrated promising potential in vision \cite{VISION_GNN_SURVEY,Graph_U_Nets,VIG,MRGCN}, owing to the intrinsic relationships present among image regions and objects. For example, ViG \cite{VIG} treats image patches as nodes and learns graph-based features for diverse downstream tasks, while DeepGCNs \cite{MRGCN} extend common vision paradigms such as residual connections and dilated convolutions to graph convolution networks (GCNs). Successful cases are also observed in few-shot segmentation (FSS), where prior methods mainly leverage graphs to enhance support–query interactions for feature alignment \cite{PPNET,SAGNN,MSGA,METAL,DAN,NIPS_SAM_GRAPH,ICCV_ONES_GRAPH}. PGNet \cite{ICCV_ONES_GRAPH} pioneers this paradigm by leveraging graph attention to propagate support label information to the query. MSGA \cite{MSGA} follows this approach, extending it with bidirectional supervision between support and query features. SAGNN \cite{SAGNN} concatenates multi-scale support–query features as nodes and mines cross-scale relationships through their interactions. PPNet \cite{PPNET} treats image patches as nodes and leverages GNNs to propagate information from unlabeled data to enhance support prototypes for improved matching. 
However, the internal relationships within images are largely ignored by these FSS methods. By contrast, our method not only leverages graph properties for support–query matching, but also exploits graphs to model the inherent and critical anatomical structures in medical images.

\section{Methodology}

\subsection{Problem Formulation}
Cross-domain few-shot medical image segmentation aims to generalize a segmentation model $\Theta$ to novel categories $\mathcal{C}_{n}$ in an unseen target domain $\mathcal{D}_t$, given only a limited number of annotated examples. The model is required to learn from a single source domain $\mathcal{D}_s$ with labeled base categories $\mathcal{C}_{b}$, and is directly evaluated on $\mathcal{D}_t$ without any retraining or fine-tuning. Notably, the source and target domains exhibit distributional shifts
and have a disjoint category set, \ie $\mathcal{C}_b \cap \mathcal{C}_n = \emptyset$.

To train our model, we randomly sample episodic tasks from the source domain $\mathcal{D}_s$ for meta-training.
Each episode $(\mathcal{S}, \mathcal{Q})$ consists of a support set $\mathcal{S} = \{(\mathbf{x}_s^i, \mathbf{y}_s^i(c))\}_{i=1}^K$ and a query set $\mathcal{Q} = \{(\mathbf{x}_q, \mathbf{y}_q(c))\}$, where $\mathbf{x}$ denotes an input image and $\mathbf{y}(c)$ is the corresponding segmentation mask for an arbitrary class $c \in \mathcal{C}_{b}$. Finally, the trained model is evaluated on test episodes sampled from  $\mathcal{D}_t$, where each query set $\mathcal{Q}$ contains only unlabeled images, and the corresponding segmentation class is drawn from  $\mathcal{C}_n$.

\subsection{Overview}
An overview of the proposed C-Graph framework is illustrated in Fig. \ref{fig:overall_arch}. 
Initially, we employ a shared-weight encoder to map input images from the image space $\mathcal{X}$ to the feature space $\mathcal{F}$, obtaining support and query features: $\mathcal{X} \to \mathcal{F}$.
Then, these features are organized into a graph space $\mathcal{Z}$, by projecting spatial pixels as nodes and defining edges via semantic affinities: $\mathcal{F} \to \mathcal{Z}$. 
Afterwards, a stack of $l$ Structural Prior Graph (SPG) layers, defined as 
$(\mathcal{Z} \times \mathcal{Y}) \times \mathcal{Z} \to \mathcal{Z} \times \mathcal{Z}$,
is hierarchically applied in a structure-to-structure manner to adapt the graph representations to the current task, and progressively model structure from local to global as domain-agnostic priors. Here, $\times$ denotes the Cartesian product, and parentheses emphasize the formation of a coupled support graph–label pair.
The output of the SPG layers is finally decoded into the label space $\mathcal{Y}$ via Subgraph Matching Decoding (SMD), which leverages node connectivity as a structural constraint for prediction: $(\mathcal{Z} \times \mathcal{Y}) \times \mathcal{Z} \to \mathcal{Y}$.
Guided by the prediction, a Confusion-minimizing Node Contrast (CNC) loss is applied to the query graph to perform semantic node contrast, thereby enhancing the discriminability of node representations in $\mathcal{Z}$.

\subsection{Modeling Image Features as a Graph.} 
In this paper, we interpret each image feature as a graph to explicitly model semantic relations between spatial locations. Specifically, let $\mathbf{F} \in \mathbb{R}^{C \times H \times W}$ denote a feature map extracted by a convolutional encoder in the feature space $\mathcal{F}$, where $C$ denotes the channel dimension, and $H$ and $W$ denote the height and width of the feature map, respectively. For each spatial location $(h, w) \in [1, H] \times [1, W]$, we define a $C$-dimensional node feature vector $\boldsymbol{\nu}_i$, indexed by $i = (h - 1) \times W + w$: 
\begin{equation}
    \boldsymbol{\nu}_i =  [\mathbf{F} + \mathbf{E}]_{:, h, w} \in \mathbb{R}^C,
\end{equation}
where $\mathbf{E} \in \mathbb{R}^{C \times H \times W}$ denotes a learnable positional encoding, and the subscript of $[\cdot]_{:,h,w}$ denotes taking all channels at $(h,w)$. This operation enables the SPG layers to incorporate spatial positional information, providing a foundation for understanding higher-level semantic relations within the modeling of structure. 
Thus, the resulting node set is given by 
    $\mathcal{V} = \left\{\, \boldsymbol{\nu}_i
    \;\middle|\; h=1,\dots,H;\; w=1,\dots,W \,\right\}$. Note that although $\mathcal{V}$ is defined as a set, its elements are consistently stacked in their original index order during the following feedforward process, \ie $\mathcal{V}$ is treated as a tensor of shape $\mathbb{R}^{C \times H \times W}$.
Correspondingly, we define each edge $\boldsymbol{e}_{i,j}$ using cosine similarity to reflect semantic relationships: 
\begin{equation}
    \boldsymbol{e}_{i,j} = \operatorname{cos}(\boldsymbol{\nu}_i, \boldsymbol{\nu}_j) = \frac{\boldsymbol{\nu}_i^\top \boldsymbol{\nu}_j}{\|\boldsymbol{\nu}_i\|_2 \cdot \|\boldsymbol{\nu}_j\|_2} \in \mathcal{E}, 
\end{equation}
where $\operatorname{cos}(\cdot, \cdot)$ denotes the cosine similarity function, and $\mathcal{E}$ denotes the edge set.

In this way,  we construct a graph $\mathcal{G} = (\mathcal{V}, \mathcal{E})$ from the image feature map to represent structure, where nodes correspond to pixel-wise features and edges reflect semantic relationships. We apply this operation to both the support and query features to construct two graph instances $\mathcal{G}_s = (\mathcal{V}_s, \mathcal{E}_s)$ and $\mathcal{G}_q = (\mathcal{V}_q, \mathcal{E}_q)$.

\subsection{Structural Prior Graph Layer}
In our proposed framework, nodes belonging to a specific category constitute a subgraph $\mathcal{G}(c) = (\mathcal{V}(c), \mathcal{E}(c))$. SPG aims to adaptively capture subgraph semantic dependencies from the support set and inject them into the query graph, and ultimately model domain-agnostic structure across the entire graph.

Technically, we leverage transformers \cite{tranformer} and graph convolution network (GCN) to model semantic relations among nodes, following the classical aggregate–update paradigm \cite{MRGCN} for iterative message passing. The SPG layer consists of three sequential stages, as illustrated below.

\noindent \textbf{Support Subgraph Linking (SSL).} 
The distribution of support subgraph nodes in the semantic space is inevitably dispersed, due to the backbone’s limited task-specific representation capacity. 
This stage enables global semantic dependency modeling among subgraph nodes, which simultaneously mitigates semantic heterogeneity and enhances subgraph compactness.

Since semantic similarity is represented as edges in the graph, our key insight is that the transformer can be interpreted as a special case of GCN, enabling global reasoning over the input graph.
Specifically, given the input $\mathbf{V}_q \in \mathbb{R}^{C \times N_q}$ as the query nodes, and $\mathbf{V}_k \in \mathbb{R}^{C \times N_k}$ serving as both the key and value nodes, the transformer computes the adjacency matrix as:
\begin{equation}
\label{eq:trans_adjmat}
\mathbf{A}_t = \operatorname{softmax} \Bigl( \tfrac{(\mathbf{V}_q^\top \mathbf{W}_q) (\mathbf{V}_k^\top \mathbf{W}_k)^{\top}}{\sqrt{C}} \Bigr) \in\mathbb{R}^{N_q \times N_k},
\end{equation}
where $\mathbf{A}_t$ denotes the adjacency matrix, $\operatorname{softmax}(\cdot)$ denotes the softmax function, and $\mathbf{W}_q , \mathbf{W}_k \in \mathbb{R}^{C \times C}$ are learnable projection matrices. 
Afterwards, node features are aggregated and updated via the computed adjacency matrix, which can be formalized as:
\begin{equation}
\label{eq:update_n_aggregate}
\begin{aligned}
\mathbf{V}_q' &\;=\; \mathcal{T}_{\theta} (\mathbf{V}_q, \mathbf{V}_k) \\
        &\;=\;
\overbrace{%
  \operatorname{FFN}\Bigl(
      \underbrace{%
        (\mathbf{A}_t \mathbf{V}_k^\top \mathbf{W}_v)^\top%
      }_{\text{Aggregate}}
      ;\,
      \mathbf{W}^o
  \Bigr)%
}^{\text{Update}}
\;\in\; \mathbb{R}^{C \times N_q},  
\end{aligned}
\end{equation}
where $\mathbf{V}_q'$ denotes the updated query graph nodes, $\mathbf{W}_v \in \mathbb{R}^{C \times C}$ denotes the projection matrix, $\operatorname{FFN}(\cdot; \mathbf{W}^o)$ denotes the feed-forward network comprising two residual connections and two layer normalizations, with $\mathbf{W}^o \in {\mathbb{R}^{C \times C}}$ as the output projection. Taken together, $\mathcal{T}_{\theta}(\cdot, \cdot)$ denotes the transformer operation, where $\theta$ comprises all learnable parameters.

We adopt a vanilla transformer to model pairwise connectivity among support subgraph nodes via self-attention:
\begin{equation}
\label{eq:mask_node}
    \mathcal{V}_s(c) = \rho(\mathcal{V}_s \odot \mathcal{R}(\mathbf{y}_s(c)), N) \in \mathbb{R}^{C \times N},
\end{equation}
\begin{equation}
    \mathcal{V}_s(c)' = \mathcal{T}_{\theta_1}(\mathcal{V}_s(c), \mathcal{V}_s(c)) \in \mathbb{R}^{C \times N},
\end{equation}
where $\mathcal{V}_s(c), \mathcal{V}_s(c)'$ denote the extracted and updated support subgraph nodes of class $c$, respectively, $\rho(f, n)$ denotes the adaptive average pooling \cite{AAP} operation that resizes the input feature map $f$ to a fixed output size $n$ along its last dimension, $\odot$ denotes the element-wise multiplication,  $\mathcal{R}(\cdot)$ resizes and broadcasts $\mathbf{y}_s(c)$ to be the same size as $\mathcal{V}_s$, \textit{i.e.}, $\mathcal{R}(\mathbf{y}_s(c)) \in ~\{0,1\}^{C \times H \times W}$, and $N$ denotes the total number of processed nodes.

\noindent \textbf{Interactive Subgraph Injection (ISI).}
The absence of query labels during inference prevents query subgraph localization, thereby hindering its subgraph structure modeling and the enforcement of connectivity constraints. In this stage, we inject structural knowledge from the support set into the query graph by leveraging the support subgraph learned previously to guide support-query alignment in the node space. Meanwhile, the query graph is updated through support-query connectivity–weighted aggregation, which highlights nodes belonging to the target category region. Specifically, we implement this operation through a cross-attention transformer, which is formalized as:
\begin{equation}
\left\{
\begin{aligned}
    \mathcal{V}_q' &= \xi^{-1}(\mathcal{T}_{\theta_2}(\xi(\mathcal{V}_q + \mathbf{I}_{pos}), \mathcal{V}_s(c)')) \\
    \mathcal{V}_s' &= \xi^{-1}(\mathcal{T}_{\theta_2}(\xi(\mathcal{V}_s + \mathbf{I}_{pos}), \mathcal{V}_s(c)'))
\end{aligned}
\right.,
\end{equation}
where $\mathcal{V}_q', \mathcal{V}_s' \in \mathbb{R}^{C \times H \times W}$ denote the updated query and support node features, respectively. 
$\xi: \mathbb{R}^{C \times H \times W} \to \mathbb{R}^{C \times HW}$ is a reshape operation with $\xi^{-1}$ as its inverse, $\mathbf{I}_{\text{pos}} \in \mathbb{R}^{C \times H \times W}$ denotes the 2D sinusoidal positional encoding used to preserve the spatial location information of graph nodes, and $\theta_2$ denotes the parameters of the cross-attention transformer. Note that the same operation is applied to the support graph to ensure consistency with the query graph in the graph space, facilitating subsequent modeling.

\noindent \textbf{Graph Structure Modeling (GSM).}
Although previous stages implicitly modeled semantic relations, explicitly capturing semantic dependencies from a spatial perspective remains essential yet underexplored in structure modeling. Moreover, graph optimization during training should extend beyond a single subgraph, enabling updates to non-target categories and thereby generalizing to unseen classes. To this end, we employ dynamic GCNs to aggregate semantic neighborhood information for each spatial node, explicitly modeling the global structure.

To begin with, considering instance-level variations within each category, we dynamically estimate explicit edges between each spatial node $\boldsymbol{\nu}_i$ and its $k$-nearest neighbors based on semantic similarity, represented by an adjacency matrix $\mathbf{A} \in \{0,1\}^{HW \times HW}$:
\begin{equation}
\label{eq:graph_adj}
\mathbf{A}(i, j) = 
\begin{cases}
1, & \text{if } \boldsymbol{\nu}_j \in \underset{\boldsymbol{\nu}_j \in \mathcal{V} \setminus \{\boldsymbol{\nu}_i\}}{\mathrm{arg\,top}_k}~ \boldsymbol{e}_{i,j}, \\
0, & \text{otherwise},
\end{cases}
\end{equation}
where $\mathbf{A}(i, j) = 1$  denotes a directed edge from  $\boldsymbol{\nu}_i$  to $\boldsymbol{\nu}_j$, $\mathrm{arg\,top}_k$ with condition $\boldsymbol{\nu}_j \in \mathcal{V} \setminus \{\boldsymbol{\nu}_i\}$ denotes the operator that returns the $k$ nodes $\boldsymbol{\nu}_j$ with the highest semantic edge weights $\boldsymbol{e}_{i,j}$. 

Subsequently, we employ max-relative graph convolution (MRConv)\cite{MRGCN} and a residual connection to update each spatial node based on the connectivity defined in $\mathbf{A}$:
\begin{equation}
\label{eq:GCN}
 \boldsymbol{\nu}_i' = \phi(\mathcal{C}(\boldsymbol{\nu}_i, \max_j \{(\boldsymbol{\nu}_i - \boldsymbol{\nu}_j) \cdot \mathbb{I}[\mathbf{A}(i,j) = 1]\}); \mathbf{W}_\phi) + \boldsymbol{\nu}_i,
\end{equation}
where $\mathbb{I}[\cdot]$ denotes an indicator function that equals 1 when the condition holds, $\mathcal{C}(\cdot, \cdot)$ denotes the concatenation operation, and $\phi(\cdot; \mathbf{W}_\phi)$ the node update function with learnable projection matrix $\mathbf{W}_\phi \in \mathbb{R}^{C \times C}$. MRConv captures semantic discontinuities and leverages directional cues in $\boldsymbol{\nu}_i - \boldsymbol{\nu}_j$ to promote node clustering, enhancing class separability in graph space.
Eq.~\ref{eq:graph_adj} and Eq.~\ref{eq:GCN} are applied to both the support and the query graph, yielding the updated graphs $\mathcal{G}_s^o = (\mathcal{V}_s^o, \mathcal{E}_s^o)$ and $\mathcal{G}_q^o= (\mathcal{V}_q^o, \mathcal{E}_q^o)$ as the final output of the SPG layer.

\subsection{Subgraph Matching Decoding}
\label{section:SMD}
Prior (CD-) FSMIS methods typically average support features into a prototype and then perform pixel-wise matching with query features via cosine similarity. However, this paradigm suffers from three key limitations: 1) Averaged prototypes fail to capture intra-class variation \cite{GMRD}; 2) Pixel-wise matching treats each query pixel vector as an isolated entity and tends to yield segmentation inconsistent with class morphology; 3) Cosine similarity uses vector direction only, leaving the attention-induced magnitude modulation in SPG layers underutilized.

To address these limitations, we propose SMD as a significant departure from the prototypical paradigm.
SMD captures semantic relations among query nodes by aggregating their connectivity to each support subgraph node. Moreover, it formulates the matching process as an attention-like computation, aligning naturally with the knowledge modeled in SPG layers. 
Specifically, we first project the nodes into a shared embedding space:
\begin{equation}
\left\{
\begin{aligned}
\widetilde{\mathcal{V}}_{s}^l(c) &= \mathbf{W}_{s}\mathcal{V}_{s}^l(c) \in \mathbb{R}^{C \times N} \\
\widetilde{\mathcal{V}}_{q}^l &= \xi(\mathcal{V}_{q}^l)^\top \mathbf{W}_{q} \in \mathbb{R}^{HW \times C}
\end{aligned}
\right.,
\end{equation}
where $\mathcal{V}_s^l$ and $\mathcal{V}_q^l$ respectively denote the support and query graph nodes from the final SPG layer. The subgraph nodes $\mathcal{V}_s^l(c)$ are extracted following a strategy similar to Eq.~\ref{eq:mask_node}.   $\widetilde{\mathcal{V}}_s^l(c)$ and $\widetilde{\mathcal{V}}_q^l$ denote the projected node features obtained via learnable linear matrices $\mathbf{W}_s$ and $\mathbf{W}_q \in \mathbb{R}^{C \times C}$.

Secondly, considering  matching needs to generalize to unseen concepts, inspired by \cite{Context_conv}, we self-update the node channel weights, as such weights are known to be task-specific \cite{channel_FSS}:
\begin{equation}
\mathcal{V}_{s}^o(c) = \bigl( 1+\tanh(\mathbf{W}_{a}\widetilde{\mathcal{V}}_{s}^l(c))\bigr)\odot \widetilde{\mathcal{V}}_{s}^l(c) \in \mathbb{R}^{C \times N},
\end{equation}
where $\mathcal{V}_{s}^o(c)$ denotes the updated support nodes, and $\mathbf{W}_{a} \in \mathbb{R}^{C \times C}$ denotes a learnable matrix.
Here, $1 + \tanh(\cdot) \in (0,2)$  highlights important channels and suppresses unimportant ones. 

Thirdly, a connectivity map $\Phi \in \mathbb{R}^{N \times H \times W}$ is computed, where each channel encodes the semantic connectivity between a support node and all query nodes, reflecting regional semantic dependencies. $\Phi$ is subsequently fused with the query nodes in a decoder $\mathcal{D}$ to aggregate semantic relations in the final node classification process:
\begin{equation}
\Phi=\sigma(\psi(\widetilde{\mathcal{V}}_q^l\mathcal{V}_{s}^o(c))),
\end{equation}
\begin{equation}
\label{eq:prediction}
\widetilde{\mathbf{y}}_q(c)
=\sigma(\mathcal D(\Phi, \mathcal{V}_q^l))
\in \mathbb{R}^{H\times W},
\end{equation}
where $\sigma(\cdot)$ denotes the sigmoid activation, $\psi: \mathbb{R}^{HW\times N} \to \mathbb{R}^{N \times H \times W}$ is a reshape function, and $\mathcal{D}$ is composed of residual blocks that produce the final prediction $\widetilde{\mathbf{y}}_q(c)$. The corresponding background prediction is computed as $\widetilde{\mathbf{y}}_q(0) = 1 - \widetilde{\mathbf{y}}_q(c)$, where $c=0$ denotes the background.

\subsection{Confusion-Minimizing Node Contrast}
To address node ambiguity and subgraph heterogeneity, we propose a contrastive strategy to mitigate semantic confusion. The core idea is to pull highly confused nodes within the target subgraph toward its semantic center, while pushing those outside the subgraph away, which enforces the learning of more semantic-oriented relationships, regardless of appearance-induced ambiguity and heterogeneity. However, the distribution of highly confused nodes is instance- and category-variant, making it impractical to localize them with a fixed pattern.
 To handle this variability, we adaptively localize such nodes based on classification entropy, which is computed from model predictions obtained in Eq.~\ref{eq:prediction}: 
\begin{equation}
    \mathbf{H} = -  \sum_{i \in \{0, c\}} \widetilde{\mathbf{y}}_q(i) \log(\widetilde{\mathbf{y}}_q(i)) \in \mathbb{R}^{H \times W},
\end{equation}
where $\mathbf{H}$ denotes the entropy map. We identify highly confused query nodes by thresholding $\mathbf{H}$, and obtain their category labels through $\mathbf{y}_q(c)$. This process is formalized as:
\begin{equation}
\label{eq:hard_mask}
   \mathbf{M}_\delta = \mathbb{I}\left[\mathbf{H} > \delta\right] \in \{0,1\}^{H \times W},
\end{equation}
\begin{equation}
\left\{
\begin{aligned}
\mathbf{p} &= \mathcal{V}_q^l \,\odot\, \mathbf{M}_\delta \,\odot\, \mathbf{y}_q{(c)} \\
\mathbf{n} &= \mathcal{V}_q^l \,\odot\, \mathbf{M}_\delta \,\odot\, \bigl(1-\mathbf{y}_q{(c)}\bigr)
\end{aligned}
\right.,
\end{equation}
where $\mathbf{M}_\delta$ denotes a spatial mask for extracting high-confusion nodes above the threshold $\delta$. $\mathbf{p}$ and $\mathbf{n}$ denote the confused nodes inside and outside the subgraph, respectively, containing $|\mathbf{p}|$ and $|\mathbf{n}|$ nodes. Only non-zero nodes after masking are retained, so $|\mathbf{p}|$ and $|\mathbf{n}|$ are instance-dependent.
Subsequently, the semantic center node $\mathbf{q} \in \mathbb{R}^{C}$ is computed by averaging the subgraph nodes,  serving as the anchor for the following semantic contrast:
\begin{equation}
    \mathbf{q} = \frac{1}{|\mathbf{y}_q(c)|}\sum_{}\mathcal{V}_q^l \odot \mathcal{R}(\mathbf{y}_q(c)).
\end{equation}

Finally, we compute the semantic edges from highly confused nodes to the semantic center as a cost matrix $\mathbf{J} \in \mathbb{R}^{|\mathbf{p}| \times (1+|\mathbf{n}|)}$:
\begin{equation}
    \mathbf J
=
\begin{pmatrix}
\operatorname{cos}(\mathbf q,\mathbf p_1)
      & \operatorname{cos}(\mathbf q,\mathbf n_1) &  \cdots  & \operatorname{cos}(\mathbf q,\mathbf n_{|\mathbf n|})\\
\operatorname{cos}(\mathbf q,\mathbf p_2)
      & \operatorname{cos}(\mathbf q,\mathbf n_1) &  \cdots  & \operatorname{cos}(\mathbf q,\mathbf n_{|\mathbf n|})\\
\vdots & \vdots & \ddots & \vdots\\
\operatorname{cos}(\mathbf q,\mathbf p_{|\mathbf p|})
      & \operatorname{cos}(\mathbf q,\mathbf n_1) &  \cdots  & \operatorname{cos}(\mathbf q,\mathbf n_{|\mathbf n|})
\end{pmatrix},
\end{equation}
where the subscripts of $\mathbf{p}$ and $\mathbf{n}$ denote the indices of the nodes.
The objective is to strengthen positive edges by increasing their confidence, while cutting off negative ones. Accordingly, the CNC loss $\mathcal L_{cnc}$ can be formalized as:
\begin{equation}
    \mathcal L_{cnc}
= -\frac1 {|\mathbf{p}|}\;
  \langle 
     \mathbbm{1},\;
     \log \bigl(
        \operatorname{softmax}(\mathbf J / \tau )
     \bigr)
  \rangle,
\end{equation}
where $\mathbbm{1}$ denotes a mask matrix indicating the first column as positive node pairs, $\tau$ denotes a temperature parameter.

\subsection{Training Objective}
We employ the standard binary cross-entropy loss $\mathcal{L}_{seg}$ to measure the distance between predictions and the corresponding ground truth, mathematically defined as:
\begin{equation}
 \mathcal{L}_{seg} =
-\frac{1}{HW} \sum_{i \in \{0, c\}}\sum_{h,w}
\mathbf{y}_{q}(i) \log \widetilde{\mathbf{y}}_q(i).
\end{equation}

Therefore, the final training loss is formulated as a weighted sum of the segmentation loss $\mathcal{L}_{\text{seg}}$ and the contrastive loss $\mathcal{L}_{\text{cnc}}$, which is formalized as:
\begin{equation}
    \mathcal{L}_{total} = \mathcal{L}_{seg} + \alpha\mathcal{L}_{cnc},
\end{equation}
where $\alpha$ is a weighting coefficient that balances the contributions of the segmentation loss and the contrastive loss. Consequently, C-Graph is supervised to model graphs with structure-informed and semantically discriminative nodes during training, thereby enabling robust and precise segmentation that generalizes well across domains.

\section{Experiments}
\subsection{Experiment Setup}
\noindent \textbf{Datasets and Protocols.}
We comprehensively evaluate our proposed model under various cross-domain scenarios, using four datasets from distinct domains:
1) \textbf{Abdominal CT} consists of 20 3D CT scans collected from the MICCAI 2015 Multi-Atlas Labeling Challenge \cite{abdct}.
2) \textbf{Abdominal MRI} includes 20 3D T2-SPIR scans from the ISBI 2019 Combined Healthy Abdominal Organ Segmentation Challenge  \cite{chaost2}.
3) \textbf{Cardiac b-SSFP} and 4) \textbf{Cardiac LGE} sourced from the MICCAI 2019 Multi-sequence Cardiac MR Segmentation Challenge \cite{cmr,CMR2019}, with 45 3D cardiac MRI scans in each dataset acquired using the b-SSFP and LGE sequences, respectively.

The abdominal datasets focus on cross-modality scenarios, where we evaluate the model on categories liver, left kidney (LK), right kidney (RK), and spleen. The cardiac datasets target cross-sequence scenarios, using classes left ventricular myocardium (LV-MYO), right ventricular myocardium (RV), and left ventricular blood pool (LV-BP) for assessment. In the following sections, we denote cross-domain directions using the symbol `$\to$', \eg CT $\to$ MRI indicates CT as the source domain and MRI as the target domain. 

\noindent \textbf{Evaluation Metric.} 
For a fair comparison \cite{FAMNet,ssl-alp2020}, we employ the Dice Sørensen coefficient (DSC) as the evaluation metric. Mathematically, DSC is defined as:
\begin{equation}
    \text{DSC}(\mathbf{y}, \widetilde{\mathbf{y}}) = \frac{2 |\mathbf{y} \cap \widetilde{\mathbf{y}}|}{|\mathbf{y}| + |\widetilde{\mathbf{y}}|} \times 100\%,
\end{equation}
where $\mathbf{y}$ denotes the ground-truth label and $\widetilde{\mathbf{y}}$ denotes the predicted segmentation result.
DSC quantifies the overlap between predicted and ground-truth labels, ranging from 0\% to 100\%. A DSC of 100\% indicates a perfect segmentation.

\begin{table*}[!t]
\vspace{-4ex}
\centering
\caption{Quantitative results (DSC \%) under cross-modality scenarios. The best results are indicated in bold.}
        \vspace{-1ex}
\resizebox{0.9\textwidth}{!}{
\begin{tabular}{ll>{\centering\arraybackslash}p{0.04\linewidth}>{\centering\arraybackslash}p{0.04\linewidth}>{\centering\arraybackslash}p{0.04\linewidth}>{\centering\arraybackslash}p{0.04\linewidth}>{\centering\arraybackslash}p{0.04\linewidth}>{\centering\arraybackslash}p{0.04\linewidth}>{\centering\arraybackslash}p{0.04\linewidth}>{\centering\arraybackslash}p{0.04\linewidth}>{\centering\arraybackslash}p{0.04\linewidth}>{\centering\arraybackslash}p{0.04\linewidth}}
\toprule
\multirow{2}{*}{Method} & \multirow{2}{*}{Reference} & \multicolumn{5}{c}{Abdominal CT $\to$ MRI} & \multicolumn{5}{c}{Abdominal MRI $\to$ CT} \\ 

\cmidrule(lr){3-7} \cmidrule(lr){8-12}

                        &                       & \multicolumn{1}{c}{Liver} & \multicolumn{1}{c}{LK} & \multicolumn{1}{c}{RK} & \multicolumn{1}{c}{Spleen} & Mean  & \multicolumn{1}{c}{Liver} & \multicolumn{1}{c}{LK} & \multicolumn{1}{c}{RK} & \multicolumn{1}{c}{Spleen} & Mean  \\ \cmidrule[1.2pt](lr){1-12}
 PANet                  & \cite{PANET}&                             39.24&                                26.47&                           37.35&         26.79&                      32.46&                        40.29&                    30.61&               26.66&   30.21& 31.94\\
 SSL-ALP                & \cite{ssl-alp}&                             70.74&                                55.49&                           67.43&         58.39&                      63.01&                        71.38&                    34.48&               32.32&   51.67&47.46\\
  ADNet                  & \cite{ADNET}                &                             50.33&                                39.36&                           37.88&         39.37&                      41.73&                        64.25&                    37.39&               25.62&   42.94& 42.55\\
 RPT                    & \cite{rpt}             &                             49.22&                                42.45&                           47.14&         48.84&                      46.91&                        65.87&                    40.07&               35.97&   51.22& 48.28\\
GMRD                    & \cite{GMRD}            &                             63.15&                                61.79&                           67.69&         56.89&                      62.38&                        66.12&                    57.38&               56.37&   54.56& 58.61\\
 PATNet                 & \cite{PATNET}               &                             57.01&                                50.23&                           53.01&         51.63&                      52.97&                        \textbf{75.94}&                    46.62&               42.68&   63.94& 57.29\\ 
  PMNet                  & \cite{PMNET}               &                            64.50&                                60.16&                           61.83&         51.80&                      59.57&                             66.82&                         39.21&                    30.87&        47.49&        46.10\\
 IFA                    & \cite{IFA}               &                             48.81&                                45.79&                           51.46&         51.42&                      49.37&                        50.05&                    36.45&               32.69&   43.08& 40.57\\ 
  APM-M& \cite{APM}&                             70.85&                                55.41&                           58.68&         53.11&                      59.51&                        74.48&                    56.01&               49.83&   64.12& 61.11\\ 
 RobustEMD& \cite{ROBUSTEMD}&                            60.16&                                66.34&                           70.26&         53.71&                      62.61&                             69.82&                         63.79&                    50.34&        59.88&        60.95\\  
  FAMNet& \cite{FAMNet}&                            \textbf{73.01}              &                                57.28&                           74.68&         58.21&                      65.79&                             73.57&                         57.79&                    61.89&        \textbf{65.78}       &        64.75\\  
 \textbf{Ours}& \textemdash    &70.92&\textbf{73.69}& \textbf{82.51}&\textbf{64.18}&\textbf{72.83}&69.60&\textbf{70.00}& \textbf{63.95}&65.23&\textbf{67.20}\\ 
 \bottomrule
\end{tabular}
}
\label{ABD}
\end{table*}

\begin{table*}[!t]
\vspace{-3ex}
\centering
\caption{Quantitative results (DSC \%) under cross-sequence scenarios. The best results are indicated in bold.}
        \vspace{-1ex}
\resizebox{0.9\textwidth}{!}{
\begin{tabular}{ll>{\centering\arraybackslash}p{0.05\linewidth}>{\centering\arraybackslash}p{0.05\linewidth}>{\centering\arraybackslash}p{0.05\linewidth}>{\centering\arraybackslash}p{0.05\linewidth}>{\centering\arraybackslash}p{0.05\linewidth}>{\centering\arraybackslash}p{0.05\linewidth}>{\centering\arraybackslash}p{0.05\linewidth}>{\centering\arraybackslash}p{0.05\linewidth}}
\toprule
\multirow{2}{*}{Method} & \multirow{2}{*}{Reference} & \multicolumn{4}{c}{Cardiac LGE $\to$ b-SSFP } & \multicolumn{4}{c}{Cardiac b-SSFP $\to$ LGE} \\ 

\cmidrule(lr){3-6} \cmidrule(lr){7-10}

                        &                       & \multicolumn{1}{c}{LV-BP} & \multicolumn{1}{c}{LV-MYO} & \multicolumn{1}{c}{RV} & Mean  & \multicolumn{1}{c}{LV-BP} & \multicolumn{1}{c}{LV-MYO} & \multicolumn{1}{c}{RV} & Mean  \\ \cmidrule[1.2pt](lr){1-10}

 PANet                  & \cite{PANET}
& 51.43& 25.75& 25.75& 36.66& 36.24                      & 26.37                       & 23.47                   & 28.69 \\
 SSL-ALP                & \cite{ssl-alp}
& 83.47& 22.73& 66.21& 57.47& 65.81& 25.64& 51.24& 47.56\\
 ADNet                  & \cite{ADNET}                
& 58.75                      & 36.94                       & 51.37                   & 49.02 & 40.36                      & 37.22                       & 43.66                   & 40.41 \\
 RPT                    & \cite{rpt}             
& 60.84                      & 42.28                       & 57.30                   & 53.47 & 50.39                      & 40.13                       & 50.50                   & 47.00 \\
 GMRD                   & \cite{GMRD}            
& 76.23& 36.87& 62.91& 58.67& 66.69& 47.19& 58.21& 57.36\\
 PATNet                 & \cite{PATNET}               
& 65.35& 50.63& 68.34& 61.44& 66.82& 53.64& 59.74& 60.06\\ 
 PMNet                  & \cite{PMNET}               
&                            73.46&                             32.11&                         68.70&       58.09&                            57.14&                             30.13&                         60.12&        49.13\\
 IFA                    & \cite{IFA}               
& 64.04& 43.22& 74.58& 62.28& 68.07& 36.07& 60.42& 54.85\\ 
 APM-M& \cite{APM}               
& 68.91& 45.74& 61.78& 58.81& 57.72& 42.37& 52.83& 50.97\\ 
 RobustEMD& \cite{ROBUSTEMD}&                            75.32&                             51.32 &                         72.86 &       66.50&                            73.19&                             50.02&                         60.29&        61.16\\ 
 FAMNet& \cite{FAMNet}&                            86.64&                             51.84&                         76.26&       71.58&                            \textbf{77.37}&                             52.05&                         54.75&        61.39\\ 
 \textbf{Ours}&\textemdash             & \textbf{87.61}& \textbf{55.22}& \textbf{79.76}& \textbf{74.20}&68.46&\textbf{56.38}&\textbf{65.14}& \textbf{63.33}\\ 
 \bottomrule
\end{tabular}}
\label{CARDIAC}
\vspace{-3ex}
\end{table*}

\noindent \textbf{Implementation Details.}
\label{implementation_details}
Our method is implemented in PyTorch \cite{pytorch}. Experiments are conducted on an NVIDIA GeForce RTX 4080 SUPER GPU with 16 GB of memory. Following the common practice in \cite{ssl-alp}, we perform data pre-processing consisting of: 1) clipping the top 5\% of intensities; 2) resampling 3D volumes to a uniform voxel spacing; and 3) center-cropping each slice to a spatial size of 256 $\times$ 256.  
As in \cite{ADNET}, supervoxel-based pseudo labels are generated to provide supervisory signals during model training. Following most of the existing FSMIS methods \cite{ADNET,qnet,rpt}, we consider the 1-way 1-shot setting in this paper. 

We employ ResNet-50 \cite{resnet} pretrained on part of MS-COCO \cite{coco} as the image encoder for our model and all compared methods.
Each image slice is repeated three times along the channel dimension to match the input format of the pretrained backbone.
The hyperparameter settings are as follows: the SPG layer depth is $l = 3$, with neighborhood size $k = 9$ used to represent a linearly varying neighborhood range across layers, from $k$ in the first layer to $2k$ in the last. The entropy threshold is $\delta = 0.2$, the temperature parameter is $\tau = 0.1$, and the loss weighting coefficient $\alpha = 0.01$. The hyperparameters in our model are determined by grid search \cite{ssl-alp} on the source domain, following the ``training-domain validation'' method in \cite{hyper_param_tuning}.

We trained our model for 48K iterations with a batch size of 1. In each iteration, a 3D volume is randomly selected from the source domain, and image–label pairs are then randomly sampled from this volume to construct the support and query sets used for training. The Adam optimizer \cite{adam} is employed for parameter updates. The learning rate is initially set as $1 \times 10^{-5}$, and decays by a factor of 0.95 every 1K iterations.
During inference, each 3D volume in the target domain is first divided into three chunks. For each chunk, we select the central slice among those containing the target class as the support image, and use the remaining slices as queries for evaluation. For each class, the mean DSC across all 3D volumes in the target domain is reported as the final result.

\begin{table*}[t]
    \vspace{-4ex}
  \centering
  \caption{Quantitative results (DSC \%) on source domains (conventional FSMIS task). The best results are indicated in bold.}
          \vspace{-1ex}
  \label{tab:SOURCE_DOMAIN}
  \resizebox{\textwidth}{!}{%
  \begin{tabular}{ll>{\centering\arraybackslash}p{0.03\linewidth}>{\centering\arraybackslash}p{0.03\linewidth}>{\centering\arraybackslash}p{0.03\linewidth}>{\centering\arraybackslash}p{0.03\linewidth}>{\centering\arraybackslash}p{0.03\linewidth}>{\centering\arraybackslash}p{0.03\linewidth}>{\centering\arraybackslash}p{0.03\linewidth}>{\centering\arraybackslash}p{0.03\linewidth}>{\centering\arraybackslash}p{0.03\linewidth}>{\centering\arraybackslash}p{0.03\linewidth}>{\centering\arraybackslash}p{0.043\linewidth}>{\centering\arraybackslash}p{0.06\linewidth}>{\centering\arraybackslash}p{0.02\linewidth}>{\centering\arraybackslash}p{0.03\linewidth}}
    \toprule
    \multirow{2}{*}{Method} & \multirow{2}{*}{Backbone}
      & \multicolumn{5}{c}{\textbf{Abdominal CT}}
      & \multicolumn{5}{c}{\textbf{Abdominal MRI}}
      & \multicolumn{4}{c}{\textbf{Cardiac b-SSFP}} \\
    \cmidrule(lr){3-7} \cmidrule(lr){8-12} \cmidrule(lr){13-16}
      & 
      & Liver & LK    & RK    & Spleen & Mean
      & Liver & LK    & RK    & Spleen & Mean
      & LV-BP & LV-MYO & RV    & Mean \\
    \cmidrule[1.2pt](lr){1-16}
    SSL-ALP  [6]& ResNet-101& 78.29& 72.36& 71.81& 70.96& \textbf{73.35}& 76.10& 81.92& 85.18& 72.18& 78.84& 83.99& 66.74& \textbf{79.96}& 76.90\\
    ADNet \cite{ADNET}           & ResNet-101&      77.24& 72.13& \textbf{79.06}& 63.48& 72.97& \textbf{82.11}& 73.86& 85.8& 72.29& 78.51& \textbf{87.53}& 62.43& 77.31& 75.76\\
     SSL-ALP \cite{ssl-alp2020}   & ResNet-50& 73.18& 70.03& 67.63& 70.04& 70.22& 73.68& 79.44& 82.69& 70.86& 76.66& 85.67& 59.44& 75.31& 73.47\\
    ADNet \cite{ADNET}           & ResNet-50&      77.54& 70.36& 75.91& 65.44& 72.31& 80.18& 72.06& 84.52& 68.59& 76.34& 81.14& 58.09& 71.42& 70.22\\
    \cmidrule[1pt](lr){1-16}
    RobustEMD \cite{ROBUSTEMD}   & ResNet-50&      \textbf{79.30}& 66.67& 54.75& 67.10& 66.96& 75.22& 68.53& 84.32& 69.78& 74.46& 80.25& 57.37& 69.41& 69.01\\
    FAMNet \cite{FAMNet}         & ResNet-50&      74.29& 71.14& 66.13& 70.08& 70.41& 80.77& 71.2& 87.21& 67.14& 76.58& 86.32& 61.98& 67.84& 72.05\\
    \textbf{Ours}                & ResNet-50&      75.89& \textbf{77.51}& 67.64& \textbf{71.35}& 73.10& 74.95& \textbf{83.48}& \textbf{88.34}& \textbf{73.44}& \textbf{80.05}& 86.48& \textbf{68.34}& 79.67& \textbf{78.16}\\
    \bottomrule
  \end{tabular}%
      }
    \vspace{-3ex}
\end{table*}

\vspace{-2ex}
\subsection{Comparison with State-of-the-Art Methods}
We compare our method with the SOTA approaches, including CD-FSMIS methods RobustEMD \cite{ROBUSTEMD} and FAMNet \cite{FAMNet}; CD-FSS methods PATNet \cite{PATNET}, PMNet \cite{PMNET},  IFA \cite{IFA} and APM-M \cite{APM}; FSMIS methods SSL-ALP \cite{ssl-alp}, ADNet \cite{ADNET}, RPT \cite{rpt}, and GMRD \cite{GMRD}; as well as our baseline method PANet \cite{PANET}. Tables \ref{ABD} and \ref{CARDIAC} present the quantitative results under cross-modality and cross-sequence scenarios, respectively. 

As shown in Table \ref{ABD}, our model surpasses FAMNet by a notable margin of 7.04\% and 2.45\% in DSC, attaining 72.83\% and 67.20\% under the Abdominal CT $\to$ MRI and MRI $\to$ CT settings, respectively. Furthermore, it significantly outperforms a broad range of CD-FSS and FSMIS approaches, with DSC improvements ranging from 25.92\% to 9.82\% for CT $\to$ MRI, and from 26.63\% to 8.59\% for MRI $\to$ CT. Although our model shows slightly inferior performance on the liver and spleen classes in the MRI $\to$ CT direction, the mean DSC still substantially exceeds FAMNet, underscoring the more comprehensive generalization ability of our approach. 

As illustrated in Table \ref{CARDIAC}, our model also achieves impressive mean DSC scores of 74.20\% and 63.33\% under the Cardiac LGE $\to$ b-SSFP and b-SSFP $\to$ LGE directions, respectively, surpassing FAMNet by 2.62\% and 1.94\%. The anatomical complexity of cardiac structures, especially the ring-shaped LV-MYO, presents significant challenges for accurate segmentation. Our model effectively captures intra-class structural dependencies via graph modeling, leading to an average improvement of 3.86\% on LV-MYO over FAMNet across both directions.

Notably, our model demonstrates superior performance in segmenting smaller anatomical structures, \ie LK, RK, and LV-MYO. Across the four directions, it yields a mean DSC improvement of 7.70\% over the second-best method on these three classes. We attribute this to the graph structure, which models class-wise features as a correlated whole and facilitates more balanced attention during training. Unlike prototypical matching methods that treat pixels as isolated entities, our model enables superior regional understanding and better preserves fine-grained structural details.

\begin{figure}[t]
    \centering
    \includegraphics[width=0.95\linewidth]{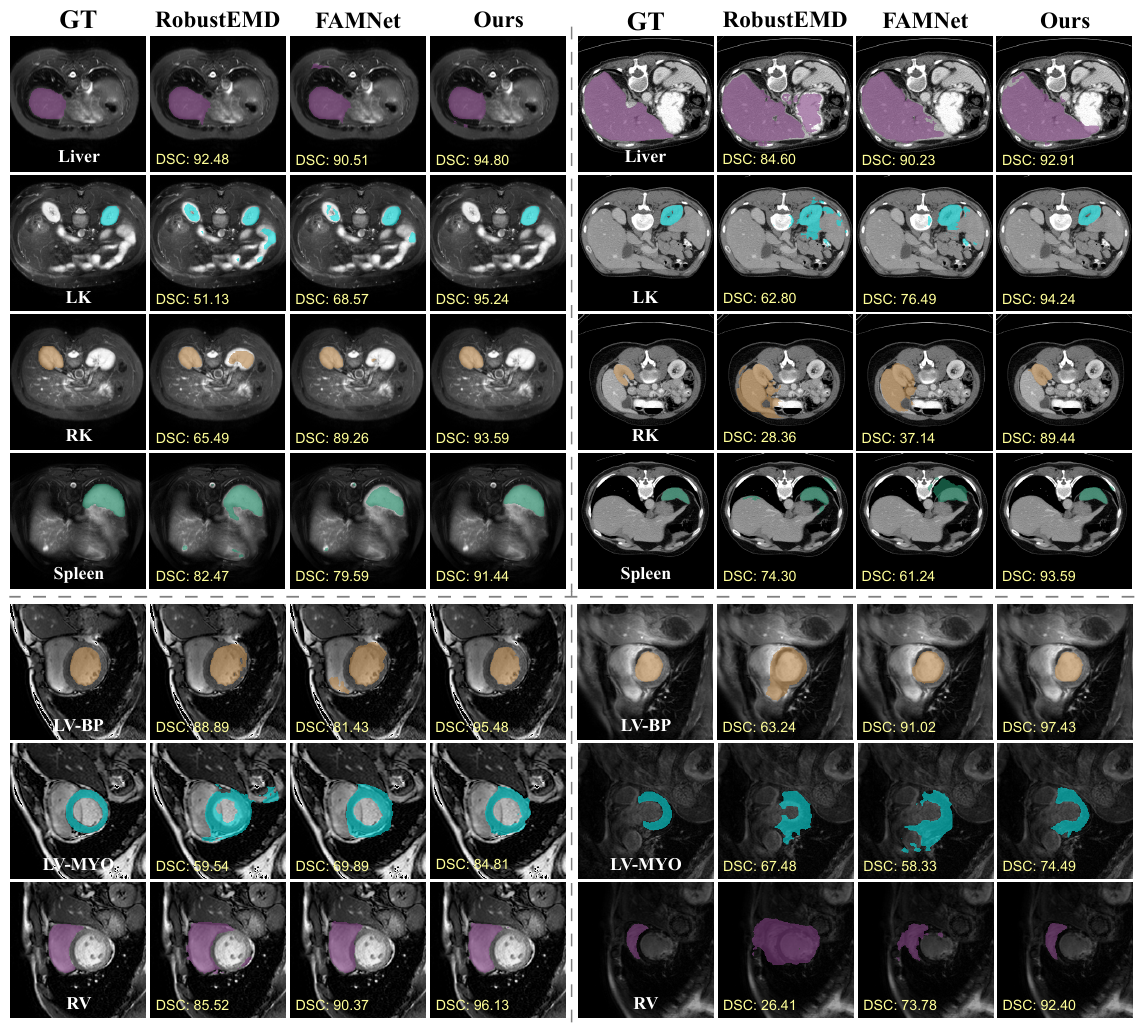}
        \vspace{-2ex}
    \caption{Visual and quantitative comparison of segmentation results. Top-left: CT $\to$ MRI; top-right: MRI $\to$ CT; bottom-left: LGE $\to$ b-SSFP; bottom-right: b-SSFP $\to$ LGE.
}
    \label{fig:segmentation}
    \vspace{-2ex}
\end{figure}

\vspace{-2ex}
\subsection{When Generalization Backfires: Source-Domain Segmentation}
Do existing CD-FSMIS methods \cite{ROBUSTEMD,FAMNet} perform well on the source domain? In this section, we investigate this question. Returning to the conventional FSMIS task \cite{ssl-alp2020}, we quantitatively evaluate the source-domain segmentation performance of different models, as illustrated in Table \ref{tab:SOURCE_DOMAIN}. 

We observe that although existing CD-FSMIS methods achieve commendable generalization to unseen domains, their source-domain performance is unexpectedly disappointing. In contrast, our method achieves results comparable to those of popular FSMIS approaches \cite{ssl-alp2020,ADNET}, and even outperforms them on abdominal MRI and cardiac b-SSFP by a large margin, despite using a backbone with weaker representational capacity \cite{rpt,PGRNET,DIFD}.
We attribute the source-domain degradation of existing CD-FSMIS methods to their common design philosophy: the suppression of domain-specific signals, \eg texture signals considered in \cite{ROBUSTEMD} and frequency components targeted in \cite{FAMNet}. While such filtering enhances domain generalizability, it compromises the model’s capacity to capture rich source-domain cues. In contrast, rather than suppressing domain-specific signals, our method exploits structural information as a bridge across domains, which enhances cross-domain transferability while preserving source-domain representability necessary for specialized performance.

\vspace{-2ex}
\subsection{Qualitative Analysis}

\subsubsection{Comparisons of Segmentation Results}
This section provides an intuitive comparison of segmentation quality between our model and existing CD-FSMIS approaches  RobustEMD \cite{ROBUSTEMD} and FAMNet \cite{FAMNet} through visualizations in Fig. \ref{fig:segmentation}. 
Our model significantly outperforms existing methods across all four cross-domain directions, particularly in preserving the structural integrity and independence of the segmented regions.
Notably, as prototypical matching-based methods, both RobustEMD and FAMNet suffer from over-segmentation when handling small classes, \eg LK, RK and LV-MYO. In contrast, our model yields segmentation results with significantly clearer inter-class distinctions and enhanced intra-class integrity. These improvements can be attributed to the SMD mechanism, which emphasizes class structural patterns, and the node contrast loss $\mathcal{L}_{cnc}$, which mitigates both intra- and extra-subgraph confusion.

\begin{figure}[t]
    \centering
    \includegraphics[width=0.95\linewidth]{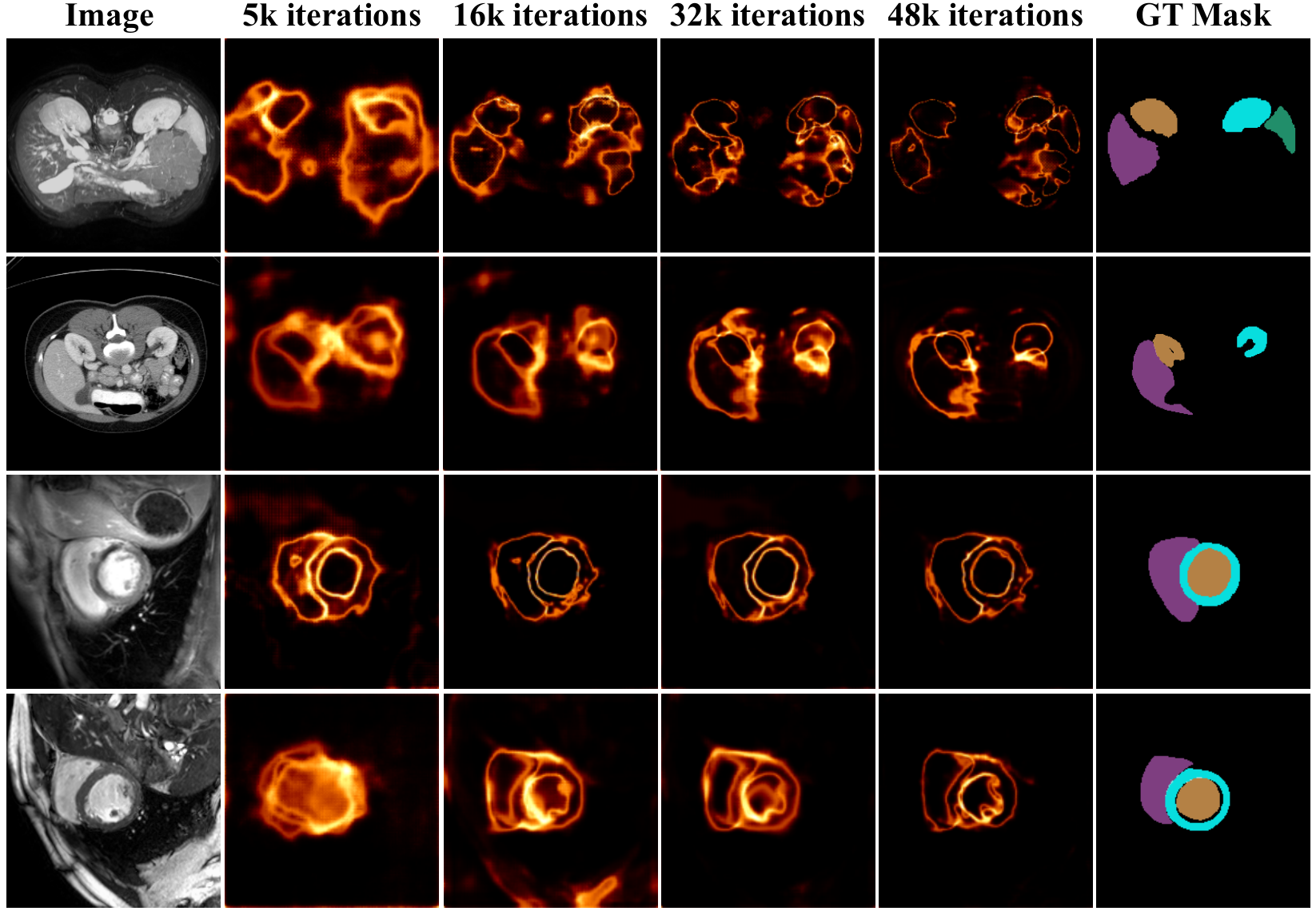}
            \vspace{-2ex}
    \caption{Iteration of entropy maps for prediction results. As training progresses, confusion decreases from high (bright yellow) to low (dark red), indicating that $\mathcal{L}_{cnc}$ significantly improves node discriminability.}
    \label{fig:confusion_iteration}
        \vspace{-2ex}
\end{figure}
\begin{figure*}[t]
    \centering
    \includegraphics[width=0.99\linewidth]{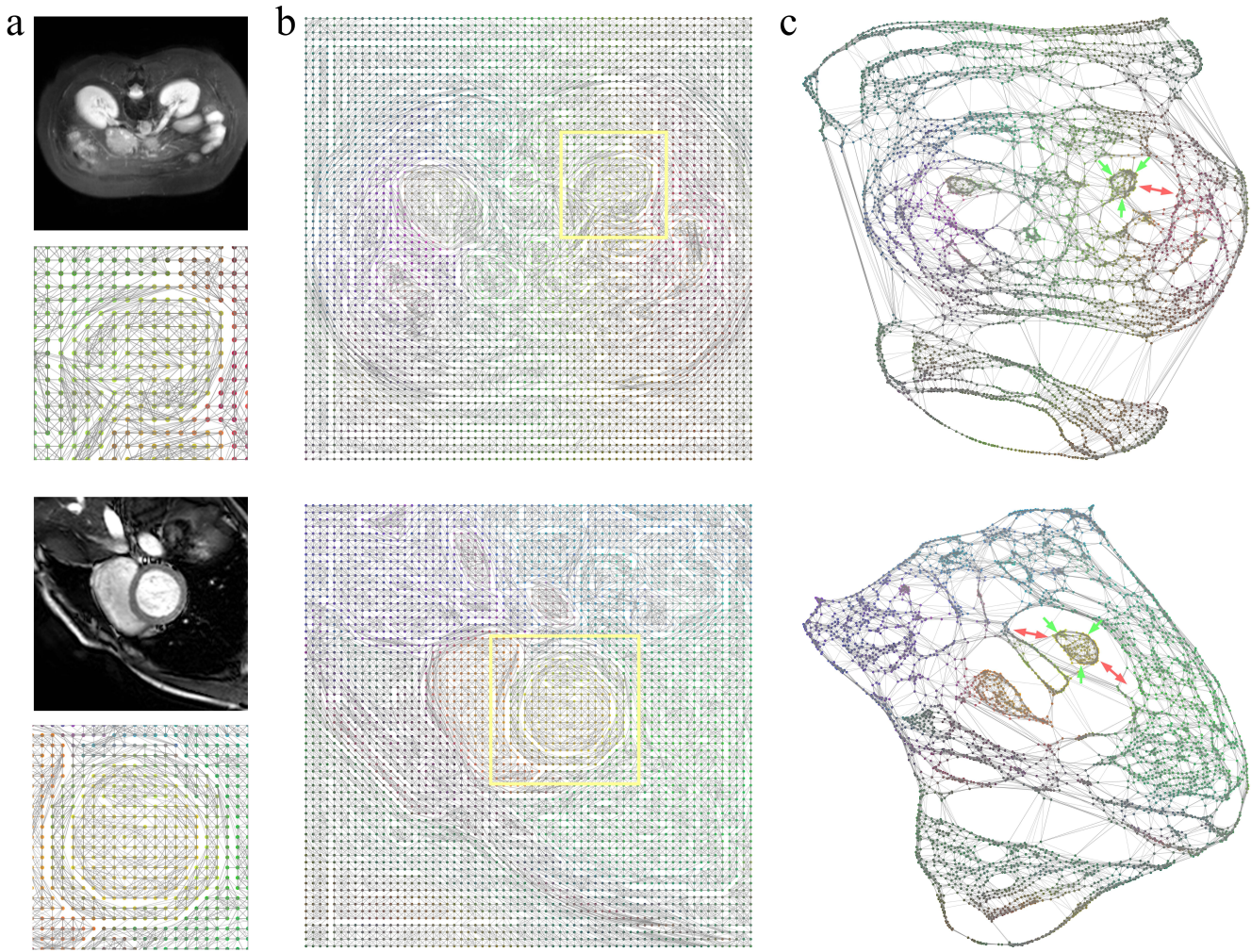}
            \vspace{-2ex}
    \caption{Visualization of the graphs constructed from intermediate features, consistent with the graph structure employed in the model pipeline. (a) Target domain images and zoomed-in view of the yellow box regions.  (b) Graph representations over feature map coordinates. (c) UMAP \cite{UMAP} visualization of the graphs in feature space. All graphs are constructed with $k=9$. Nodes with the same color denote the same entity across the graphs in (b) and (c).
The edges collectively reflect anatomical semantics in the target-domain image, indicating that the graph effectively captures its underlying structural patterns. Best viewed in color and with zoom.}
    \label{fig:visual_graph}
    \vspace{-3ex}
\end{figure*}

\subsubsection{Analysis of Confusion Minimization}
In this section, we qualitatively analyze our model’s contrastive learning ability in minimizing node semantic confusion.
Fig.~\ref{fig:confusion_iteration} visualizes the confusion maps of the model at different training iterations. As training progresses, the model shows a clear improvement in distinguishing high-confusion regions. 
The penultimate column (48K iterations) presents the final confusion map. Notably, the design of $\mathcal{L}_{cnc}$ emphasizes not only the highly confused regions near category boundaries, but also those within and around category interiors. Consequently, compared to early training stages, the final model significantly reduces confusion both within category regions and along their boundaries.

\begin{figure*}[t]
    \centering
    \includegraphics[width=0.99\linewidth]{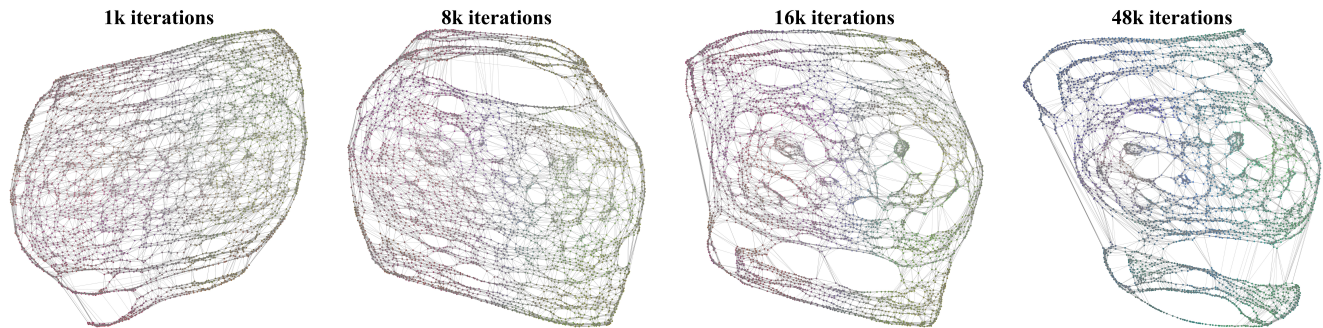}
    \vspace{-1ex}
    \caption{Evolution of estimated graph structure over training iterations. The corresponding original image is shown in the upper part of Fig. \ref{fig:visual_graph}(a). All graphs are constructed with $k=9$. Nodes with the same color (assigned via PCA projection \cite{PCA}) denote the same entity across the graphs. As training progresses, our model learns improved intra-subgraph compactness and greater inter-subgraph separability.}
    \label{fig:graph_evolution}
\end{figure*}

\begin{table}[t]
\centering
\caption{Ablation studies (DSC \%) for component effectiveness.}
\resizebox{0.95\linewidth}{!}{
\begin{tabular}{ccc>{\centering\arraybackslash}p{0.05\linewidth}c c c c c}
\toprule
\multicolumn{3}{c}{SPG Layer} & \multirow{2}{*}{\(\mathcal{L}_{cnc}\)} & \multicolumn{5}{c}{Abdominal CT $\to$ MRI} \\
\cmidrule(lr){1-3} \cmidrule(lr){5-9}
SSL & ISI & GSM &  & Liver & LK & RK & Spleen & Mean \\
\cmidrule[1.2pt](lr){1-9}
&        &        & \multirow{5}{*}{w/o}& 39.24 & 26.47 & 37.35 & 26.79 & 32.46 \\
&        & \checkmark &                   & 68.42& 49.85& 52.65& 57.22& 57.04\\
&        \checkmark & \checkmark &                   & 72.75& 67.65& 77.83& 58.69& 69.23\\
\checkmark & \checkmark & &     & 59.67& 66.57& 78.66& 65.28& 67.55\\
\checkmark & \checkmark & \checkmark &     & 72.59& 67.40 & 78.37 & 63.22 & 70.40 \\
\cmidrule[1pt]{1-9}
& &  \checkmark & \multirow{4}{*}{w/}& 69.45 & 56.92 & 60.11 & 54.94 & 60.36 \\
& \checkmark & \checkmark &            & 70.26 & 70.10 & 77.34 & 65.42 & 70.78 \\
\checkmark & \checkmark & &            & 63.37& 69.02& 80.72& 65.14& 69.56\\
\checkmark & \checkmark & \checkmark &            & 70.92 & 73.69 & 82.51 & 64.18 & 72.83 \\
\bottomrule
\end{tabular}
}
\label{tab:module_effectiveness}
\vspace{-3ex}
\end{table}
\begin{table}[t]
\centering
\caption{Quantitative comparison of different matching strategies in terms of DSC (\%) on Abdominal CT $\to$ MRI.}
\label{tab:different_matching}
\begin{tabular}{lccccc}
\toprule
Method & Liver& LK& RK& Spleen& Mean \\
\midrule
Proto. w/o Thres. & 57.19 & 59.71 & 65.23 & 58.93 & 60.27 \\
Proto. w/ Thres. & 71.69 & 70.78 & 80.93 & 62.46 & 71.39 \\
SMD w/ Pool.     & 71.25& 71.48& 76.55& 60.95& 70.06\\
SMD w/o Pool.     & 70.92& 73.69 & 82.51 & 64.18 & 72.83 \\
\bottomrule
\end{tabular}
\begin{tablenotes}
\item Here, ``Proto.'' denotes prediction obtained through prototypical
matching, ``Thres.'' represents ``Threshold'', and ``Pool." stands for ``Pooling''.
\end{tablenotes}
\label{SMD}
\vspace{-3ex}
\end{table}

\subsubsection{Graph Representation over Feature Space}
In this section, we present an interesting phenomenon by visualizing the intermediate-layer graph, as shown in Fig. \ref{fig:visual_graph}. Specifically, we visualize the explicit edges on the feature map coordinates (column b), and depict the structural relationships in the feature space using UMAP \cite{UMAP} (column c). Visualizations are shown for abdominal CT $\to$ MRI and cardiac LGE $\to$ bSSFP. Notably, the graph edges consistently and accurately reflect the underlying anatomical structures in the image, exhibiting dense intra-class connections and well-delineated inter-class boundaries, as detailed in column a. 
This observation provides compelling evidence for our Definition 1, \ie semantic relationships among spatial positions encode anatomical information. Moreover, the UMAP visualization further highlights the effectiveness of our contrastive graph modeling: nodes from the same category form highly compact clusters (green arrow), while remaining clearly separated from others (red arrow). The evolution of this property is illustrated in Fig. \ref{fig:graph_evolution}. As training progresses, the discriminability among different subgraph regions is progressively enhanced, providing a well-structured and easily separable latent graph space for subsequent model decision. These findings collectively validate our method’s ability to learn and model domain-generalizable and category-discriminative graph representations.

\vspace{-2ex}
\subsection{Component Effectiveness Analysis}
Table \ref{tab:module_effectiveness} summarizes the contribution of each component. All variants use SMD for matching. Even when trained without CNC loss, when combined with SMD, SPG layers boost performance by a substantial 37.94\% on top of the baseline (PANet, row 1), highlighting the effectiveness of structure modeling. Within SPG, GSM alone improves the baseline by 24.58\% by enabling global anatomical structure modeling beyond training categories; SSL and ISI together provide an additional 13.36\% gain, where ISI contributes the most by transferring support category knowledge, while SSL reinforces node semantic dependency learning. When employing CNC loss in training, it consistently boosts performance across various SPG variants, yielding an additional 2.43\% gain with the complete SPG and reducing the performance gap among the variants by enhancing node discriminability through contrastive learning.

\begin{table}[t]
\centering
\vspace{-4ex}
\caption{Quantitative results (DSC \%) under the 5-shot setting. The relative improvements over the 1-shot setting ($\uparrow\!\!\!$) are shown in the last column.}
\label{tab:5-shot}
\resizebox{0.95\linewidth}{!}{
\begin{tabular}{lccccc}
\toprule
Abodominal       & Liver & LK     & RK    & Spleen & Mean  \\ \midrule
CT $\to$ MRI     & 75.28 & 79.71  & 84.37 & 74.41  & 78.44 \textsubscript{$\uparrow$ 5.61} \\
MRI $\to$ CT     & 74.33& 76.42& 72.19& 69.91& 73.21 \textsubscript{$\uparrow$ 6.01}\\ \cmidrule[1pt]{1-6}
Cardiac          & LV-BP& LV-MYO & RV    &        \textemdash& Mean  \\ \midrule
LGE $\to$ b-SSFP & 90.60 & 62.50  & 85.23 &        \textemdash& 79.44 \textsubscript{$\uparrow$ 5.24} \\
b-SSFP $\to$ LGE & 73.13 & 59.37  & 72.03 &        \textemdash& 68.18 \textsubscript{$\uparrow$ 4.85} \\ \bottomrule
\end{tabular}
}
\vspace{-2ex}
\end{table}

\begin{table}[t]
\centering
\vspace{-4ex}
\caption{Quantitative results (DSC \%) under cross-context scenarios. Here, all the methods are evaluated using the model trained on Abdominal CT, and the best results are indicated in bold.}
\label{tab:cross_context}
\resizebox{0.95\linewidth}{!}{
\begin{tabular}{lccccccc}
\toprule
\multirow{3}{*}{Method}       & \textbf{Chest} & \textbf{Skin} & \multicolumn{4}{c}{\textbf{Cardiac}} \\ 
                              & \textbf{X-ray} & \textbf{Dermoscopy} & \multicolumn{4}{c}{\textbf{b-SSFP MRI}} \\  \cmidrule(lr){2-2} \cmidrule(lr){3-3} \cmidrule(lr){4-7}
                              & Lung & Lesions & LV-BP & LV-MYO & RV & Mean \\ \cmidrule[1.3pt](lr){1-7}
PANet~\cite{PANET}   & 68.02 & 35.53 & 29.69 & 19.10 & 26.78 & 25.19 \\
SSL-ALP~\cite{ssl-alp} & 71.66 & 39.16 & 61.18 & 27.14 & 48.45 & 45.59 \\
ADNet~\cite{ADNET}   & 40.32 & 22.11 & 46.61 & 28.47 & 42.93 & 39.34 \\
RPT~\cite{rpt}       & 58.48 & 35.44 & 69.35 & 42.79 & 53.11 & 55.08 \\
GMRD~\cite{GMRD}     & 52.31 & 39.66 & 41.45 & 28.36 & 38.59 & 36.13 \\
PATNet~\cite{PATNET} & 77.65 & 42.88 & \textbf{69.88} & 46.09 & 52.97 & 56.31 \\
PMNet~\cite{PMNET} & 75.65&  37.28&  62.42&  21.34&  57.93&  47.23  \\
IFA~\cite{IFA}       & 76.21 & 42.59 & 66.42 & 27.89 & 39.78 & 44.70 \\
APM-M~\cite{APM} & 71.49 & 39.85 & 44.30 & 31.17 & 45.18 & 40.22 \\
RobustEMD~\cite{ROBUSTEMD}  & 63.99  & 46.33 & 37.77 & 41.44 & 42.48 & 40.56 \\
FAMNet~\cite{FAMNet} & 67.02 & 36.45 & 58.44 & 33.00 & 51.34 & 47.59 \\
\textbf{C-Graph (Ours)} & \textbf{78.38} & \textbf{50.42} & 63.70 & \textbf{46.32} & \textbf{66.00} & \textbf{58.67} \\
\bottomrule
\end{tabular}
}
\vspace{-2ex}
\end{table}

\vspace{-2ex}
\subsection{From Prototypical Matching to SMD} 
This section investigates the effect of replacing prototypical matching with our proposed SMD. We compare two prior prototypical variants, one with learnable thresholds \cite{ADNET} and one without \cite{PANET}. We also evaluate an extreme variant of SMD in which the support graph is reduced to a single node via global average pooling (GAP), effectively approximating a prototypical matching scenario. Table \ref{SMD} reports the quantitative comparisons. SMD with complete support subgraphs surpasses the thresholded and plain prototypical baselines by 1.44\% and 12.56\%, respectively. This gain stems from the preservation of semantic relations in the subgraph for matching, which enables the structural priors learned by SPG layers to be fully leveraged.
Additionally, the SMD variant with GAP yields a 2.77\% drop in DSC compared to its unpooled counterpart, underscoring the importance of preserving accurate local connectivity within the query subgraph. Without pooling, SMD retains fine-grained correspondences between support subgraph nodes and the query graph, which we argue also helps mitigate the impact of subgraph heterogeneity.

\vspace{-1ex}
\subsection{Performance Scalability with More Shots}
This section evaluates the scalability of our model when multiple support samples ($K$-shot) are available. For evaluation, we directly use the model trained under the 1-shot setting, and support subgraphs are averaged for matching. The original support sample is retained. For the additional samples, we select $K-1$ volumes with IDs greater than the original support volume ID (modulo the dataset size) from the target domain dataset, and obtain support samples as detailed in Section \ref{implementation_details}.

As shown in Table \ref{tab:5-shot}, increasing $K$ to 5 consistently yields a performance gain of approximately 5\% over the 1-shot setting. We find that this improvement is more pronounced in categories that were relatively poorly segmented under the 1-shot scenario, such as liver and LV-MYO. We attribute this to their large inter-patient variability in appearance. With more shots, the support representation becomes less biased by individual cases and better approximates the general class distribution, thereby providing more reliable guidance for segmentation.

\begin{figure}[t]
    \centering
    \includegraphics[width=0.95\linewidth]{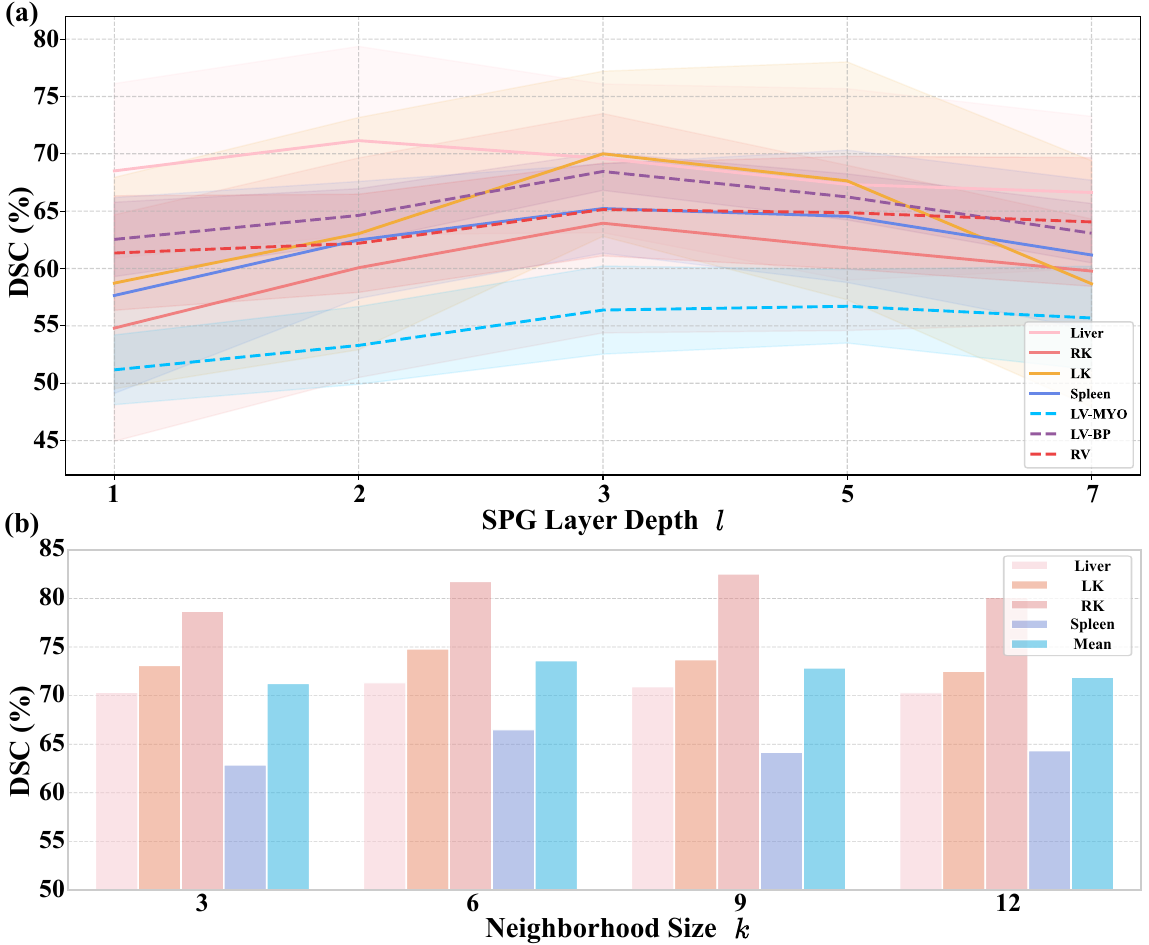}
            \vspace{-2ex}
    \caption{Hyperparameter analysis for model configuration.
(a) SPG layer depth $l$; results are reported under the abdominal MRI $\to$ CT and cardiac b-SSFP $\to$ LGE directions, with shaded areas indicating standard deviation. 
(b) Neighborhood size $k$; results are reported under abdominal CT $\to$ MRI. 
}
    \label{fig:model_config}
    \vspace{-2ex}
\end{figure}

\vspace{-1ex}
\subsection{Towards Broader Medical Domains}
In this section, we investigate a more challenging yet practical cross-domain scenario in medical imaging, where models are required to generalize across domains involving different anatomical contexts (\eg training on abdominal images and inferring on chest radiographs), a setting we refer to as cross-context. For evaluation, we use abdominal CT as the source domain and assess its performance on cardiac b-SSFP and two additional target-domain datasets: (i) Chest X-ray \cite{Chestxray1,Chestxray2}, which comprises 566 radiographs with the segmentation target being the lung, and (ii) Skin Dermoscopy \cite{isic1,isic2}, which contains 2,594 images for segmenting three lesion types (519 melanoma, 1,867 nevus, and 208 seborrheic keratosis). All images are normalized and resized to $256 \times 256$.

 As shown in Table \ref{tab:cross_context}, even without tailored designs, our model still outperforms a wide range of methods by a remarkable margin. We attribute this to its strong capability of dynamically estimating the graph structure from arbitrary images, rather than overfitting to the source-domain structure. Consequently, our model generalizes not only to domains with structures consistent with those seen during training but also to domains exhibiting substantial contextual shift.

\subsection{Hyperparameter Studies} 
\subsubsection{Model Configuration}
\noindent \textbf{SPG layer depth $l$}. Fig.~\ref{fig:model_config}(a) shows that, for most classes, accuracy peaks at a moderate depth, typically around 3 layers. The overall trend indicates a steady improvement in performance followed by a slight decline as the depth continues to increase. Notably, smaller and structurally simpler classes, \eg LK, RK, and LV-BP, are more sensitive to the number of layers.
We hypothesize that deeper SPG stacks enable the capture of longer-range structural dependencies. However, when the range becomes excessively large, the information aggregation of distant nodes may cause a blurring effect, which diminishes the semantics of class-specific nodes. On the other hand, a moderate layer depth encourages dominant intra-class aggregation while maintaining beneficial inter-class correlations, leading to improved performance with greater efficiency.

\noindent \textbf{Neighborhood size $k$} plays an important role in edge estimation, as a larger hop size allows each node to aggregate information from a broader spatial region. Experimental results in Fig.~\ref{fig:model_config}(b) show that performance initially improves and then degrades as $k$ increases, peaking at $k = 6$ with a mean DSC of 73.59\%. This trend aligns with the performance pattern observed when varying SPG layer depth $l$, as both effectively modify the receptive field of each node.

\begin{figure}[t]
    \centering
    \includegraphics[width=0.95\linewidth]{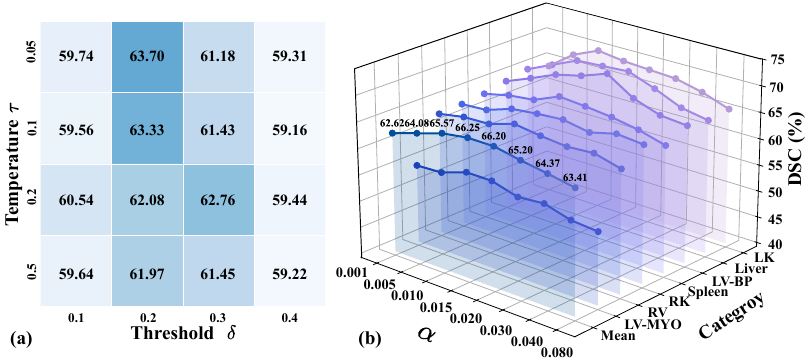}
            \vspace{-2ex}
    \caption{Hyperparameter analysis for $\mathcal{L}_{cnc}$. 
(a) Temperature parameter $\tau$ and entropy threshold $\delta$ in $\mathcal{L}_{cnc}$; results (DSC \%) are reported under cardiac b-SSFP $\to$ LGE. 
(b) Weighting Coefficient $\alpha$; results are reported under the abdominal MRI $\to$ CT and cardiac b-SSFP $\to$ LGE.
}
    \label{fig:CNC_loss_config}
    \vspace{-1ex}
\end{figure}
\begin{table}[t]
\centering
\caption{Quantitative comparison of different entropy thresholding strategies in terms of DSC (\%).}
\resizebox{0.9\linewidth}{!}{
\begin{tabular}{l>{\centering\arraybackslash}p{0.05\linewidth}c c c c c}
\toprule
\multirow{2}{*}{Threshold $\delta$} & \multicolumn{5}{c}{Abdominal CT $\to$ MRI} \\
\cmidrule(lr){2-6}
 & Liver & LK & RK & Spleen & Mean \\
\cmidrule[1.2pt](lr){1-6}
 Fixed ($\delta=0.2$)&70.92  &73.69  &\textbf{82.51}  &64.18  &72.83 \\
 Learnable (global) &70.01	&72.26	&82.32	&\textbf{68.06}	&73.16 \\
 Learnable (instance)&  \textbf{72.24}& \textbf{75.63} & 79.86 & 65.93 & \textbf{73.42}  \\
\bottomrule
\end{tabular}
}
\label{tab:learnable_thres}
\vspace{-3ex}
\end{table}

\subsubsection{Confusion-minimizing Node Contrast Loss}
\noindent \textbf{Temperature parameter $\tau$ and entropy threshold $\delta$}. 
$\tau$ controls the sharpness of edge weights in node contrast; smaller values yield stronger discrimination between nodes. $\delta$ determines the confusion level of selected nodes, with higher $\delta$ including more confused nodes. 
As shown in Fig.~\ref{fig:CNC_loss_config}(a), performance peaks at $\tau = 0.05$ and $\delta = 0.2$, achieving a mean DSC of 63.70\%. Both overly low (\eg $\delta = 0.1$) and high (\eg $\delta = 0.4$) thresholds significantly degrade the effectiveness of $\mathcal{L}_{\text{cnc}}$. We attribute this to small $\delta$ mistakenly including confident nodes, weakening the contrastive focus on truly confused nodes, while large $\delta$ yields too few samples for effective training. 
The impact of $\tau$ is relatively moderate. Generally, smaller values (\eg $\tau = 0.05$ or $\tau = 0.1$) lead to better performance, as they encourage stronger category nodes separation.

We further consider a learnable $\delta$ to adaptively sample highly confused nodes. Two methods are explored: (i) a global threshold for all entropy maps, or (ii) an instance-adaptive threshold predicted from the final SPG output query graph via a fully connected layer. To ensure gradient flow and avoid trivial solutions, inspired by \cite{gumbel_softmax}, we replace the binary mask in Eq. \ref{eq:hard_mask} with a soft weighting mask. Table \ref{tab:learnable_thres} shows that learnable $\delta$ consistently improves performance, with the instance-adaptive variant performing best by mitigating the impact of instance-specific variations, particularly for the liver, which varies greatly in appearance across individuals and compared to other organs.

\noindent \textbf{Weighting Coefficient $\alpha$} balances the contribution of $\mathcal{L}_{seg}$ and $\mathcal{L}_{cnc}$. As shown in Fig.~\ref{fig:CNC_loss_config}(b), with the increase of $\alpha$, the performance shows an increasing trend, as the model is encouraged to learn a more discriminative latent graph space to reduce segmentation confusion. However, when $\alpha$ becomes excessively large, the performance gradually decreases, which we attribute to insufficient optimization for the segmentation task. Overall, except for extremely small values of $\alpha$, where node contrast has almost no contribution, introducing node contrast consistently leads to significant improvements, indicating that $\mathcal{L}_{cnc}$ effectively complements segmentation and is highly compatible with the segmentation objective.

Overall, our method is largely insensitive to hyperparameter settings. As shown in Fig.~\ref{fig:model_config} and Fig.~\ref{fig:CNC_loss_config}, the model exhibits minimal performance fluctuation and a wide tolerance to hyperparameter variations, while still consistently surpassing other methods in Table \ref{ABD} and Table \ref{CARDIAC} across most parameter settings. These findings further demonstrate the robustness of our approach.

\begin{table}[t]
\centering
\vspace{-2ex}
\caption{Quantitative comparison of performance–efficiency trade-offs for different methods.}
\label{tab:efficiency}
\resizebox{0.95\linewidth}{!}{
\begin{tabular}{lccccc}
\toprule
\multirow{3}{*}{Method} & \multicolumn{2}{c}{\textbf{Train}} & \multicolumn{2}{c}{\textbf{Test}} & \multirow{3}{*}{mDSC}  \\ \cmidrule(lr){2-3} \cmidrule(lr){4-5}
                        & Latency      & Memory     & Latency      & Memory    &                        \\
                        & \scriptsize (ms/img)     & \scriptsize (GB)       & \scriptsize (ms/img)     & \scriptsize (GB)      &                        \\ \cmidrule[1.3pt](lr){1-6} RPT \cite{rpt}&              101.18&            5.74&              37.44&           0.49&                        62.16\\

                         GMRD \cite{GMRD}&              121.21&            7.31&              51.24&           2.11&                        68.22\\
                        
                        RobustEMD \cite{ROBUSTEMD}&              186.11&            5.95&              107.7&           1.37&                        65.95\\
                        FAMNet \cite{FAMNet}&              112.93&            1.28&              16.07&           0.42&                        68.94\\
                        \textbf{Ours}&              154.38&            5.87&              84.31&           1.04&                        72.69\\ \bottomrule
\end{tabular}
}
\begin{tablenotes}
\item Here, ``Memory'' reports the peak memory usage, and ``mDSC'' denotes the mean DSC (\%) across all tasks, including both cross-domain and source-domain results. All experiments use the same 4080 SUPER GPU.
\end{tablenotes}
\vspace{-3ex}
\end{table}

\begin{table}[t]
\centering
\caption{Quantitative analysis of the importance of domain-specific information for source-domain segmentation. The performance (DSC \%) is measured on Abdominal MRI.}
\label{tab:Domain-specific}
\resizebox{0.95\linewidth}{!}{
\begin{tabular}{lcccc}
\toprule
Method & Suppression & Performance & SSIM & $\mathcal{L}_s$ \\ \midrule
FAMNet \cite{FAMNet} &   \ding{51}    &       76.58&      0.794&          0.287\\
FAMNet \cite{FAMNet} &   \ding{55}    &       78.91&      0.822&          0.234\\
\textbf{Ours}                 &   \textemdash  &       80.05&      0.840&          0.221\\ \bottomrule
\end{tabular}
}
\vspace{-2ex}
\end{table}

\subsection{Performance–Efficiency Trade-off}
From Table \ref{tab:efficiency}, we conclude that our method achieves the highest overall performance (mDSC = 72.69\%) while maintaining efficient segmentation, thus demonstrating a superior trade-off. Compared with the best competitors, although GMRD \cite{GMRD} has relatively good performance, it consumes the highest GPU memory in both training and inference. FAMNet \cite{FAMNet} is the most efficient model in terms of latency and memory, but its overall performance falls far behind ours due to its limited performance on the source domain. In contrast, our method achieves reasonable latency and memory usage with substantially higher performance than all other approaches. This indicates that our method combines universal segmentation capability with friendly computational cost for current hardware.

\subsection{On Preserving Domain-Specific Information}
Filtering domain-specific information has been argued to be crucial to source-domain performance. To validate this, we modify FAMNet \cite{FAMNet} by reversing its suppression branch to also promote domain-specific features. We then reconstruct images from the learned features using a U-Net \cite{Unet} decoder, and evaluate domain-specific content using style difference $\mathcal{L}_s$ \cite{STYLE_LOSS} (style) and SSIM \cite{SSIM} (contrast, brightness). A smaller $\mathcal{L}_s$ or a larger SSIM score indicates that domain-specific information is better preserved in the features. Table \ref{tab:Domain-specific} shows that reversing the suppression operation (indicated by `\ding{55}') improves source-domain performance by 2.33\% DSC, accompanied by higher SSIM and lower $\mathcal{L}_s$, indicating better preservation of domain-specific information. Moreover, C-Graph achieves the best SSIM and $\mathcal{L}_s$, aligning with its superior source-domain performance. These consistent results across metrics confirm that domain-specific information is crucial for source-domain performance, indicating that its removal is inadvisable for universal segmentation.

\begin{figure}[t]
    \centering
    \includegraphics[width=0.99\linewidth]{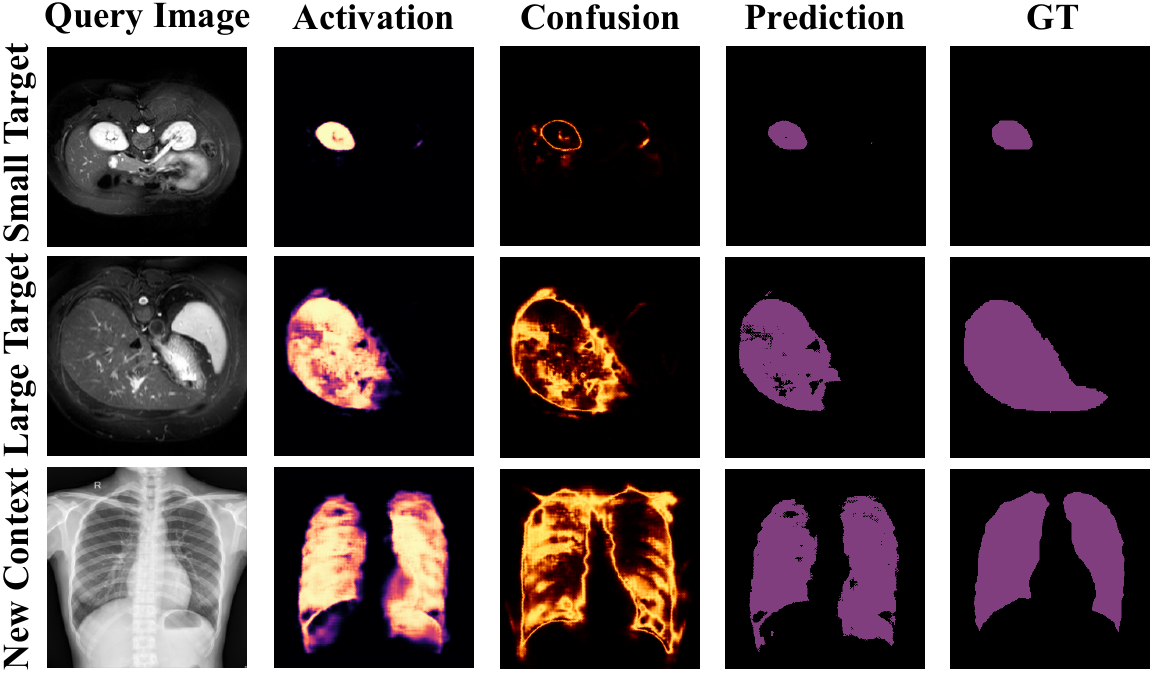}
    \caption{Case study with model trained on Abdominal CT. As shown in the first row, our method demonstrates strong performance on small targets, achieving consistently high activations over the foreground with minimal segmentation confusion. Nevertheless, in some failure cases (rows 2 and 3), the model struggles with large targets and under contextual shifts, leading to higher confusion within the foreground and its surrounding regions, along with false activations.}
    \label{fig:failure_cases}
        \vspace{-2ex}
\end{figure}

\section{Limitations and Future Work}
 Firstly, although our model achieves overwhelming superiority on smaller classes, its performance on larger ones (\eg liver) is relatively mediocre, as exemplified in Fig. \ref{fig:failure_cases}. We attribute this to the substantial class size imbalance inherent in medical images, which leads to uneven subgraph modeling performance across classes. While the relatively small neighborhood size $k$ used in our model configuration is well-suited for most classes, it limits the receptive field of graph convolution for large classes, hindering effective large-class subgraph modeling. However, we believe this limitation can be alleviated by adopting an adaptive strategy for setting $k$. 
 
Furthermore, although our model demonstrates strong performance against imaging technique shifts, its generalization is less satisfactory when the domain gap goes beyond imaging factors, such as in cross-context scenarios, resulting in a few failure cases as exemplified in Fig. \ref{fig:failure_cases}. Future research could explore a generic approach to address more diverse domain shifts in medical imaging, thereby further improving generalizability and clinical applicability.

\section{Conclusion}
We have presented C-Graph, a novel framework that leverages structural consistency in medical images for CD-FSMIS. Its core component, the SPG layer, modeled image features as graphs to capture and transfer target-category information while encoding domain-agnostic structural priors. Based on SPG outputs, we proposed a novel SMD mechanism that incorporates semantic relations among category nodes to guide prediction. Furthermore, we introduced a CNC loss to reduce semantic ambiguity and subgraph heterogeneity among nodes, thereby enhancing node discriminability in the graph space. Extensive experiments and analyses have validated the superior segmentation performance of our method in both in-domain and cross-domain scenarios. We believe our contributions will drive the next stage of CD-FSMIS development.

\section*{Acknowledgments}
The authors would like to thank the reviewers and editors for their constructive comments and valuable suggestions, which helped improve the quality of this paper.

\bibliographystyle{IEEEtran_custom}
\bibliography{reference}

\end{document}